\documentclass[letterpaper, 10 pt, journal, twoside]{IEEEtran} 

\usepackage{graphics} 
\usepackage{epsfig} 
\usepackage{mathptmx} 
\usepackage{times} 
\usepackage{amsmath} 
\usepackage{amssymb}  
\usepackage{cite}
\usepackage{color}
\usepackage{bm}
\usepackage{array}
\usepackage[symbol]{footmisc}
\usepackage[colorlinks,linkcolor=black,anchorcolor=black,citecolor=black,urlcolor=black]{hyperref}
\usepackage{tikz}

\usepackage[left=0.57in, right=0.57in, top=0.83in, bottom=1.2in]{geometry}

\graphicspath{{./images/}}

\newcommand{\cornerwidth}{0.25\textwidth}
\newcommand{\comparewidth}{0.22\textwidth}
\newcommand{\kernelwidth}{0.25\textwidth}

\newcommand\copyrighttext{%
	\scriptsize \textcopyright \centering 2019 IEEE.  Personal use of this material is permitted.  Permission from IEEE must be obtained for all other uses, in any current or future media, including reprinting/republishing this material for advertising or promotional purposes, creating new collective works, for resale or redistribution to servers or lists, or reuse of any copyrighted component of this work in other works.}
\newcommand\copyrightnotice{%
	\begin{tikzpicture}[remember picture,overlay]
	\node[anchor=south,yshift=30pt] at (current page.south) 
	{{\parbox{\dimexpr\textwidth-50\fboxsep-\fboxrule\relax}{\copyrighttext}}};
	\end{tikzpicture}%
}

\hypersetup{
	pdftitle={Asynchronous Spatial Image Convolutions for Event Cameras},
	pdfsubject={Computer Vision, Robotics},
	pdfauthor={Cedric Scheerlinck, Nick Barnes, Robert Mahony},
	pdfkeywords={Event cameras, Dynamic Vision Sensor, Convolutions, Corner detection, Low latency}
}

\author{Cedric Scheerlinck, Nick Barnes, Robert Mahony
\thanks{ \newline \indent Manuscript received September 10, 2018; accepted December 27, 2018.
	Date of publication January 16, 2019; date of current version February 8, 2019.
	This letter was recommended for publication by Associate Editor P. Vasseur and Editor E. Marchand upon evaluation of the reviewers' comments.
	This work was supported in part by the Australian Government Research Training Program
	Scholarship and in part by the Australian Research Council through the ``Australian Centre of Excellence for Robotic Vision'' under Grant CE140100016.
	
	C. Scheerlinck and R. Mahony are with The Australian National University, Canberra, ACT 0200, Australia.
	
	N. Barnes is with The Australian National University, Canberra, ACT 0200,
	Australia, and also with Data61, CSIRO, Canberra, ACT 2601, Australia.

	Digital Object Identifier \href{https://doi.org/10.1109/LRA.2019.2893427}{10.1109/LRA.2019.2893427} }
}

\title{Asynchronous Spatial Image Convolutions for \\ Event Cameras}

\begin{document}
\maketitle
\copyrightnotice
%
\begin{abstract}
Spatial convolution is arguably the most fundamental of 2D image processing operations.
Conventional spatial image convolution can only be applied to a conventional image, that is, an array of pixel values (or similar image representation) that are associated with a single instant in time.
Event cameras have serial, asynchronous output with no natural notion of an image frame, and each event arrives with a different timestamp.
In this paper, we propose a method to compute the convolution of a linear spatial kernel with the output of an event camera.
The approach operates on the event stream output of the camera directly without synthesising pseudo-image frames as is common in the literature.
The key idea is the introduction of an internal state that directly encodes the convolved image information, which is updated asynchronously as each event arrives from the camera.
The state can be read-off as-often-as and whenever required for use in higher level vision algorithms for real-time robotic systems.
We demonstrate the application of our method to corner detection, providing an implementation of a Harris corner-response ``state'' that can be used in real-time for feature detection and tracking on robotic systems.
\end{abstract}

\begin{IEEEkeywords}
Computer vision for automation, visual tracking.
\end{IEEEkeywords}

\noindent
\textbf{Website:} \url{https://cedric-scheerlinck.github.io/event-convolutions}

\section{INTRODUCTION}

\IEEEPARstart{S}PATIAL image convolutions are a core pre-processing step in almost all robotic vision algorithms.
For example, Gaussian smoothing, gradient computation, computation of the Laplacian, etc, are convolutional operations that underlie fundamental vision algorithms such as: feature detection, optical flow computation, edge detection, etc.
Classical image convolution requires a full image frame such as are generated by conventional synchronous cameras.
Event cameras \cite{Lichtsteiner08ssc,Brandli14ssc} in contrast, provide asynchronous, data-driven measurements \vspace{2cm} of grey-scale temporal contrast\footnote[2]{We consider temporal contrast events (not grey-level events \cite{Posch11ssc,Huang17iscas}).}
at high temporal resolution and dynamic range.
Event cameras have the potential to overcome many inherent limitations that conventional cameras display for robotic applications: motion blur in high-speed environments, under/overexposure in high-dynamic-range scenes,
sparse temporal sampling (low frame rate), or very high bandwidth and data requirements (high frame rate).
With such advantages, event cameras are an ideal embedded visual sensor modality for robotic systems \cite{Rebecq17ral,Mueggler15rss,Mueggler14iros,Censi14icra,Rosinol18ral}.
However, the lack of a conventional image frame means that any image processing algorithm that relies on convolution cannot be directly applied to the output of an event camera.

In this paper, we propose a novel algorithm to compute the convolution of a linear kernel with the underlying radiometric scene information encoded by the output of an event camera.
The key contribution of the paper is the introduction of an internal `state' that encodes the convolved image information.
Each pixel of the internal state carries a timestamp of the last event that updated that pixel (analogous to the surface of active events \cite{Benosman14tnnls}), along with the latest state information, for example it could be values of: horizontal and vertical gradient, the Laplacian, or a Gaussian filtered intensity, etc.

The proposed algorithm uses continuous-time filter theory to compute a filtered or time-averaged version of the input event stream.
Since spatial convolution is a linear process, it can be factored through the linear filter equations and applied directly to the event stream inputs.
Thus, each event is spatially convolved with a linear kernel to generate a neighbouring collection of events, all with the same timestamp, which are then fed into pixel-by-pixel single-input-single-output continuous-time linear filters.
The resulting filter equations can be solved explicitly, allowing asynchronous, discrete implementation of the continuous-time filter based on exact interpolation.
Each asynchronous update of the internal state requires computation of one scalar exponential along with a small number of simple algebraic operations.
The resulting algorithm does not require a motion-model for the camera and is truly asynchronous and highly efficient.
Our method does not require reconstruction of pseudo-images, avoiding the latency and computational cost associated with synchronous reconstruction.
The internal state can be separately read-off as-often-as and whenever required by a separate processing thread for use in higher level vision algorithms.

\newgeometry{left=0.57in, right=0.57in, top=0.67in, bottom=0.65in}

We demonstrate our approach using a variety of common kernels including Gaussian, Sobel and Laplacian kernels.
To provide a more substantial example, we apply the method to the estimation of Harris corners.
The approach taken is to augment the internal linear filter state with a (non-linear) Harris corner-response state.
This `state' is computed from the various gradients asynchronously as they are updated and provides a real-time measure of the Harris corner response of the underlying radiometric scene.
The Harris corner state provides estimates of corners that we compare to a frame-based Harris detector, as well as state-of-the-art event-based corner detectors.
We emphasise that in our algorithm no grey-scale image was required, or indeed is generated.

The remainder of the paper is as follows:
Section \ref{sec:related} reviews related works.
Section \ref{sec:method} outlines mathematical formulation and methodology.
Section \ref{sec:results} presents experimental results and analysis.
Section \ref{sec:conclusion} concludes the paper.

\section{Related Works}\label{sec:related}

Event cameras such as the DVS128 \cite{Lichtsteiner08ssc} and DAVIS240 \cite{Brandli14ssc} provide asynchronous, data-driven contrast events, and are popular among roboticists due to their high temporal resolution and dynamic range, and low bandwidth and power consumption.
One approach to computing spatial image convolutions with event cameras is to reconstruct image intensity \cite{Reinbacher16bmvc,Barua16wcav,Bardow16cvpr,Brandli14iscas,Scheerlinck18accv,Pan18arxiv} and apply 2D spatial convolutions to the output.
\cite{Reinbacher16bmvc,Barua16wcav,Bardow16cvpr} convert event-streams into image frames by taking either fixed temporal-windows of events, or a fixed number of events per frame.
Scheerlinck et al. \cite{Scheerlinck18accv} introduce the concept of a continuous-time image state that encodes image intensity, and is updated asynchronously with each event.
The approaches of \cite{Brandli14iscas,Scheerlinck18accv,Pan18arxiv} combine image frames with events to estimate image intensity that is available at the same temporal resolution as events.
Computing convolutions in this manner, however, is unattractive as it incurs additional computational cost in the original image reconstruction as well as introducing noise, and imposes latency as the user has to `wait' for the frame.

Event-driven gradient maps have been explored from a SLAM perspective \cite{Cook11ijcnn,Kim14bmvc,Kim16eccv} where the aim is to simultaneously estimate pose and a map.
SLAM methods require pose estimation and depend also on a motion model for the camera.
Moreover, to apply image convolution, the gradient output of these algorithms  must be converted to intensity images (e.g. via Poisson solvers \cite{Agrawal05iccv,Agrawal06eccv}).

Alternative representations for event data such as surface of active events \cite{Benosman14tnnls} or exponentially decaying time-surfaces \cite{Lagorce16pami,Sironi18cvpr}, warped-event counts \cite{Stoffregen17acra,Gallego18cvpr,Gallego17ral,Zhu17icra}, and plane-fitting \cite{Benosman14tnnls,Clady15nn} have also proved useful for tasks such as recognition, motion estimation, and feature detection and tracking.
Event-based corner detection algorithms \cite{Mueggler17bmvc,Vasco16iros,Clady15nn,Alzugaray18ral,Glover17iros} have been proposed that aim to detect corners on some form of time-surface \cite{Mueggler17bmvc,Vasco16iros,Alzugaray18ral} or intersections of planes fitted to events in space-time \cite{Clady15nn}.
These methods represent state-of-the-art in event-based corner detection, though they are not designed for generalised spatial image convolutions.

Event cameras such as ATIS \cite{Posch11ssc}, and \cite{Huang17iscas} are capable of providing absolute intensity with each event.
Ieng et al. \cite{Ieng14ieee} propose a method to asynchronously compute spatial convolutions that relies on grey-level events (provided by the ATIS \cite{Posch11ssc}), and show that beyond 3 frames per second, asynchronous convolutions outperform frame-based in computational cost.
Sabatier et al. \cite{Sabatier17tip} perform an asynchronous Fourier transform with the ATIS, demonstrating improved computational efficiency compared to frame-based methods.
Huang et al. \cite{Huang18iscas} propose an on-chip module that causes events to trigger neighbouring pixels for gradient computation.
These methods rely on event cameras that are able to provide grey-level events.
An alternative approach to event-based, 2D image filtering is the VLSI architecture proposed in \cite{serrano1999aer}, capable of implementing any convolutional kernel that is decomposable into $x$ and $y$ components.

\section{METHOD} \label{sec:method}

The proposed  method is formulated as a parallel collection of continuous-time filters that are solved asynchronously as discrete updates using exact interpolation.
That is we compute the exact analytic solution to the associated ordinary differential equation of the filter in continuous time and evaluate at discrete time instances.

\subsection{Mathematical Representation and Notation}

Each pixel in the event camera responds independently and asynchronously to changes in brightness.
When the change in log intensity relative to the previous reference level exceeds a preset threshold $c$,
\begin{align}
|\log(I) - \log(I_\text{ref})| > c,
\end{align}
an event is triggered and the pixel reference $I_\text{ref}$ resets to the new brightness level.
For contrast event cameras \cite{Lichtsteiner08ssc,Brandli14ssc}, each event contains the time-stamp ($t$; relative to a global clock), discrete pixel address \mbox{$\bm{p} = (x, y)^T$}, and polarity ($\sigma = \pm 1$ depending on the sign of the brightness change).
\begin{align}
\text{event}_i = (t_i, \bm{p}_i, \sigma_i), \quad i \in 1, 2, 3 ...
\end{align}
The output of an event camera is a serial stream of asynchronous events.

Events can be modelled as Dirac-delta functions \cite{Mueggler17ijrr}.
Define an event $e_i(\bm{p}, t)$ as
\begin{align} \label{eq:e_i}
e_i(\bm{p}, t) := \sigma_i \, c \, \delta(t - t_i) \, \delta_{\bm{p}_i}(\bm{p}),
\end{align}
where $\delta(t)$ is a Dirac-delta function and $\delta_{\bm{p}_i}(\bm{p})$ is a Kronecker delta function with indices associated with the pixel coordinates of $\bm{p}_i$ and $\bm{p}$.
That is \mbox{$\delta_{\bm{p}_i}(\bm{p}) = 1$} when \mbox{$\bm{p} = \bm{p}_i$} and zero otherwise.
In this paper we use the common assumption that the contrast threshold $c$ is constant \cite{Reinbacher16bmvc,Kim16eccv,Kim14bmvc}, although, in practice it does vary somewhat with intensity, event-rate and other factors \cite{Brandli14iscas}.
The integral of events is
\begin{align} \label{eq:L}
\int_{0}^t \sum_i e_i(\bm{p}, \tau) d \tau = L(\bm{p}, t) - L(\bm{p}, 0) + \int_{0}^t \eta(\bm{p}, \tau) d \tau,
\end{align}
where $L(\bm{p}, t)$ is the log intensity seen by the camera with initial condition $L(\bm{p}, 0)$, and $\eta(\bm{p}, t)$ represents quantisation and sensor noise. $L(\bm{p}, 0)$ is typically unknown and $\eta(\bm{p}, t)$ is unknown and poorly characterised.
If left unchecked, integrated error arising from $\int_0^t \eta(\bm{p}, \tau) d \tau$ grows over time and quickly degrades the estimate of $L(\bm{p}, t)$ \cite{Brandli14iscas}.
A method to deal with error arising from $L(\bm{p}, 0)$ and $\eta(\bm{p}, t)$ will be presented in \mbox{section \ref{sub:ct-filter}}.

\subsection{Event Convolutions}

Let $K$ denote a linear spatial kernel with finite support.
Consider the convolution of $K$ with $L(\bm{p}, t)$.
Define 
\begin{align}
L^K(\bm{p}, t) := (K * L) (\bm{p}, t).
\end{align}
Using \eqref{eq:e_i}, \eqref{eq:L} and omitting the noise term $\eta(\bm{p}, t)$ in the approximation
\begin{align}
L^K(\bm{p}, t) & \approx (K * L)(\bm{p}, 0) + \int_{0}^t \sum_i (K * e_i)(\bm{p}, \tau) d \tau, \notag \\
& \approx (K * L)(\bm{p}, 0) +   \int_{0}^t \sum_i  \sigma_i \, c \, \delta(t - t_i) \, (K * \delta_{\bm{p}_i})(\bm{p}) d \tau, \notag  \\
& \approx (K * L)(\bm{p}, 0) +   \int_{0}^t \sum_i e^K_i(\bm{p}, \tau) d \tau, \label{eq:KstarL}
\end{align}
where
\begin{align}
e^K_i(\bm{p}, t) :=  \sigma_i \, c \, \delta(t - t_i) \, (K * \delta_{\bm{p}_i})(\bm{p}). \label{eq:e^K_i}
\end{align}
Note that $(K * \delta_{\bm{p}_i})(\bm{p})$ is a local spatial convolution of the finite support kernel $K$ with a single non-zero image pixel.
The result of such a convolution is an image with pixel values of zero everywhere except for a patch centred on $\bm{p}_i$ (the same size as $K$) with values drawn from the coefficients of $K$.
The convolved event $e^K_i(\bm{p}, t)$ can be thought of as a finite (localised) collection of spatially separate events all occurring at the same time $t_i$.

\subsection{Continuous-time Filter for Convolved Events} \label{sub:ct-filter}

It is possible to compute the direct integral \eqref{eq:KstarL} using a similar approach to the direct integration schemes of \cite{Brandli14iscas,Reinbacher16bmvc}.
The drawback of this approach is integration of sensor noise, which results in drift, and undermines low temporal-frequency components of the estimate $L^K(\bm{p}, t)$ over time.
Furthermore, we are often concerned with high temporal-frequency information (i.e. scene dynamics), especially in robotic systems scenarios where the scene around the robot is changing continually.
This leads us to consider a simple high-pass filtered version of $L^K(\bm{p},t)$.

\emph{Frequency domain:}
We design the high-pass filter in the frequency domain, and later implement it in the time domain via inverse Laplace transform. For $\alpha > 0$, a scalar constant, we define a high pass filter $F(s) := s/(s+ \alpha)$ and apply it directly to the integrated event stream \eqref{eq:KstarL}.
Let $\mathcal{L}^K(\bm{p}, s)$ denote the Laplace transform of the signal $L^K(\bm{p}, t)$.
Let $\hat{\mathcal{G}} (\bm{p}, s)$ denote the high-pass filtered version of $\mathcal{L}^K (\bm{p}, s)$.
That is:
\begin{align}
\hat{\mathcal{G}} (\bm{p}, s) & := \frac{s}{s+ \alpha} \mathcal{L}^K(\bm{p},s), \notag \\
& = \frac{s}{s+ \alpha} \frac{1}{s} \sum_i \mathcal{E}^K_i (\bm{p},s)
+ \frac{s}{s+ \alpha}\frac{1}{s} (K * L)(\bm{p}, 0), \notag \\
& = \frac{1}{s+ \alpha} \sum_i \mathcal{E}^K_i (\bm{p},s)
+ \frac{1}{s+ \alpha} (K * L)(\bm{p}, 0),
\label{eq:g_hat_frequency}
\end{align}
where $\mathcal{E}^K_i (\bm{p},s) = \sigma_i \, c \, \exp(- t_i s) (K * \delta_{\bm{p}_i})(\bm{p})$ is the Laplace transform of $e^K_i (\bm{p},t)$.
The DC term associated with the unknown initial condition has an exponentially decreasing time-response $e^{-\alpha t} (K * L)(\bm{p}, 0)$ in the filter state and is quickly attenuated.
The high-pass filter naturally attenuates low-frequency components of the noise signal $\eta(\bm{p}, t)$.

\emph{Time domain:}
Ignoring the $(K * L)(\bm{p}, 0)$ initial condition, the time domain signal $\hat{G} (\bm{p}, t)$ can be computed by taking the inverse Laplace of \eqref{eq:g_hat_frequency} and solving the resulting ordinary differential equation\footnote[8]{Although we write this as a partial differential equation (the partial taken with respect to time) there is no coupling between pixel locations and the solution decouples into parallel pixel-by-pixel ODEs.}
\begin{align} \label{eq:high-pass}
\frac{\partial}{\partial t} \hat{G}(\bm{p}, t) = - \alpha \hat{G}(\bm{p}, t) + \sum_i e_i^K(\bm{p}, t),
\end{align}
for each pixel $\bm{p}$.
Here, $\hat{G}(\bm{p}, t)$ is a pixel-by-pixel internal state that provides an estimate of the high-pass component of \mbox{$(K*L)(\bm{p},t)$}.

The continuous-time differential equation  \eqref{eq:high-pass} is a constant coefficient linear differential equation except at time instances when an event occurs and can be solved explicitly.
To exploit this property we store the timestamp of the latest event at each pixel $t^{\bm{p}}$ and use the explicit solution of \eqref{eq:high-pass} to asynchronously update the state when (and only when) a new event at that pixel occurs.

The constant-coefficient, first-order ODE for \eqref{eq:high-pass} assuming no events is
\begin{align} \label{eq:ccode}
\frac{\partial}{\partial t} \hat{G}(\bm{p}, t) &= - \alpha \hat{G}(\bm{p}, t).
\end{align}
Let $t_i$ denote the timestamp of the current event and denote the limit to $t_i$ from below by $t_i^-$ and the limit to $t_i$ from above by $t_i^+$.
Integrate \eqref{eq:ccode} from $t^{\bm{p}}$ (the timestamp of the previous event at $\bm{p}$) to $t_i^-$
\begin{align}
\hat{G}(\bm{p}, t_i^-) &= \exp( - \alpha (t_i - t^{\bm{p}})) \hat{G}(\bm{p}, t^{\bm{p}}).
\end{align}
Next integrate \eqref{eq:high-pass} over the convolved event, i.e. from $t_i^-$ to $t_i^+$
\begin{align*}
\int_{t_i^-}^{t_i^+} \frac{\partial}{\partial t} \hat{G}(\bm{p}, t) d t &= \int_{t_i^-}^{t_i^+} - \alpha \hat{G}(\bm{p}, t) + e_i^K(\bm{p}, t) d t.
\end{align*}
The integral of the right-hand side is  $\sigma_i \, c \, (K * \delta_{\bm{p}_i})(\bm{p})$ since $\hat{G}(\bm{p}, t)$ is continuous and its infinitesimal integral is zero, and the Dirac delta integrates to unity.
Thus, one has
\begin{align}
\hat{G}(\bm{p}, t_i^+) &= \hat{G}(\bm{p}, t_i^-) + \sigma_i \, c \, (K * \delta_{\bm{p}_i})(\bm{p}). \label{eq:add_event}
\end{align}
In addition, it is necessary to update the timestamp state
\begin{align}
t^{\bm{p}} = t_i. \label{eq:timestamp}
\end{align}

Equations \eqref{eq:ccode}, \eqref{eq:add_event} and \eqref{eq:timestamp} together define an asynchronous distributed update that can be applied pixel-by-pixel to compute the filter state.
The state could also be updated at any user-chosen time-instance (for example just before a read-out) with the time-instance stored in $t^{\bm{p}}$.

Multiple different filters can be run in parallel.  For example, if gradient estimation is required, then two filter states ($\hat{G}_x, \hat{G}_y$) can be run in parallel for the $x$ and $y$ components using an appropriate directional kernels (Sobel, central difference, etc).

\section{RESULTS} \label{sec:results}
The experiments were performed using a DAVIS240C \cite{Brandli14ssc} with default biases provided in the jAER software, and sequences from \cite{Scheerlinck18accv} and \cite{Mueggler17ijrr}.
The internal filter state of the system is asynchronous and for visualisation we display instantaneous snap shots taken at sample times.
There is only a single parameter $\alpha$ in the filter.
We set $\alpha = 2 \pi\,$rad/s for all sequences unless otherwise stated.
The complexity of our algorithm scales linearly with the number of (non-zero) elements in the kernel, and we find that a kernel size of $3 \times 3$ is usually sufficient. We fix the contrast threshold $c$ constant.

\subsection{Event Convolutions}
\begin{figure*}[t]
	\centering
	\resizebox{0.77\textwidth}{!}{\begin{tabular}{ >{\centering\arraybackslash} m{2.5cm} >{\centering\arraybackslash} m{4.5cm} >{\centering\arraybackslash} m{4.5cm} >{\centering\arraybackslash} m{4.5cm}}
		& \texttt{snowman} & \texttt{sun} & \texttt{night\_drive}
		\\
		Identity
		
		\medskip
		$
		\begin{bmatrix}
		0 & 0 & 0 \\
		0 & 1 & 0 \\
		0 & 0 & 0
		\end{bmatrix}
		$ &
		\includegraphics[width=\kernelwidth]{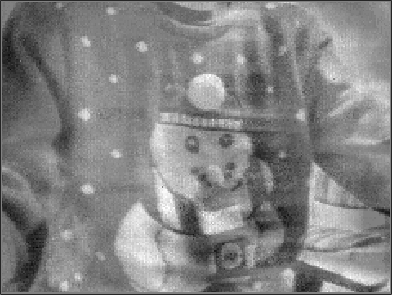}
		&
		\includegraphics[width=\kernelwidth]{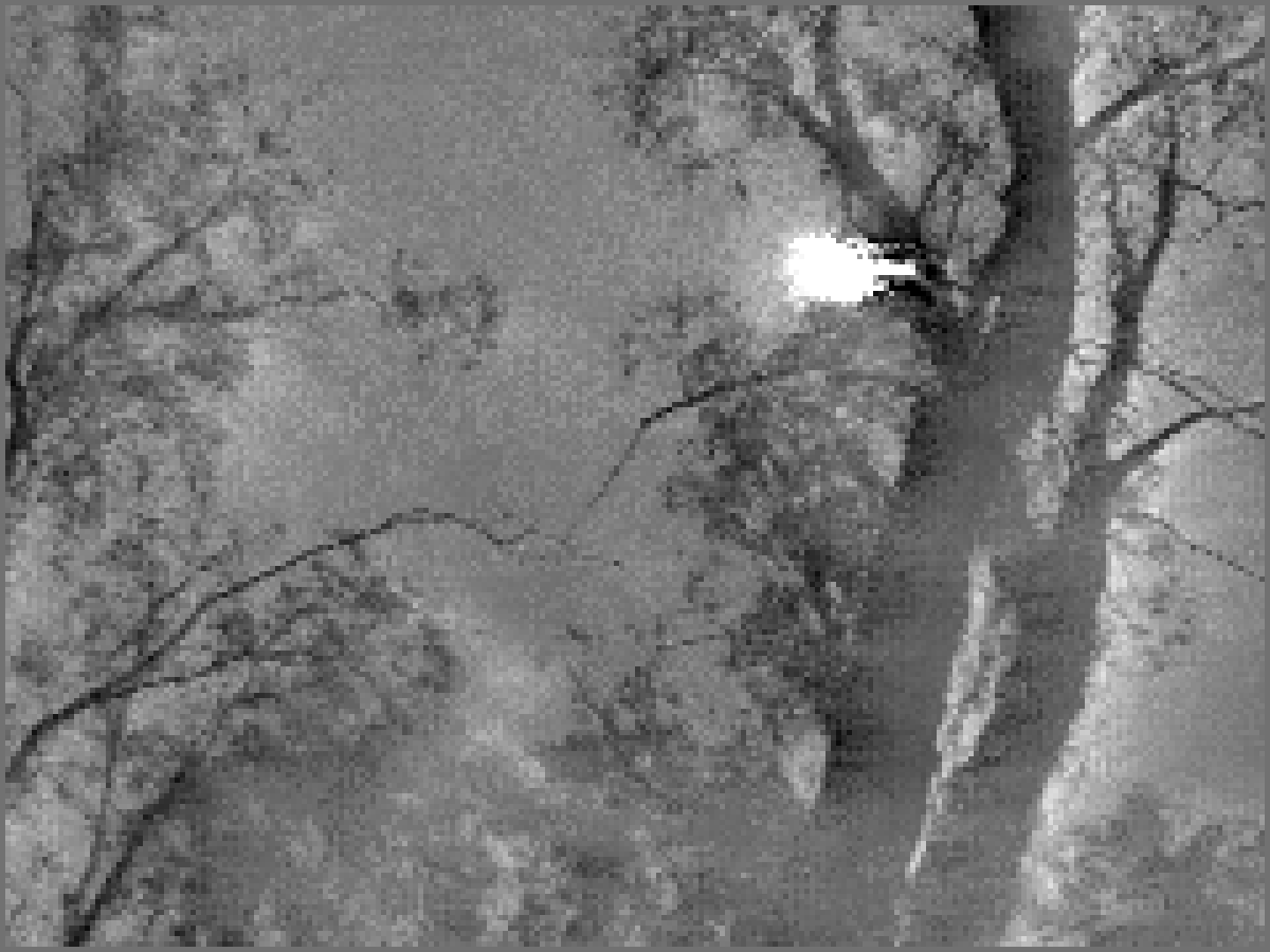}
		&
		\includegraphics[width=\kernelwidth]{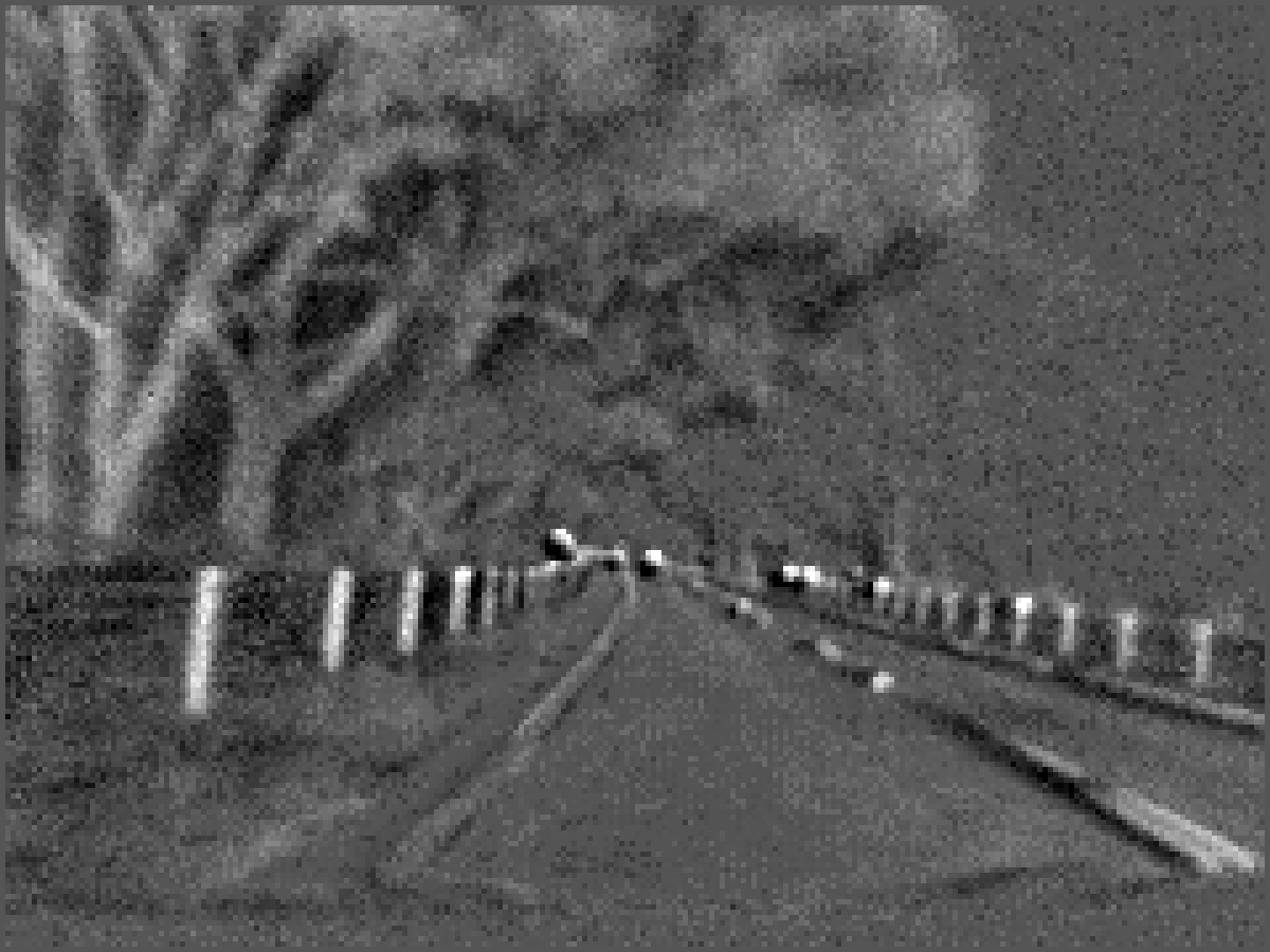}
		\\
		Gaussian
		
		\medskip
		$
		5 \times 5, \  \sigma = 3.0
		$ &
		\includegraphics[width=\kernelwidth]{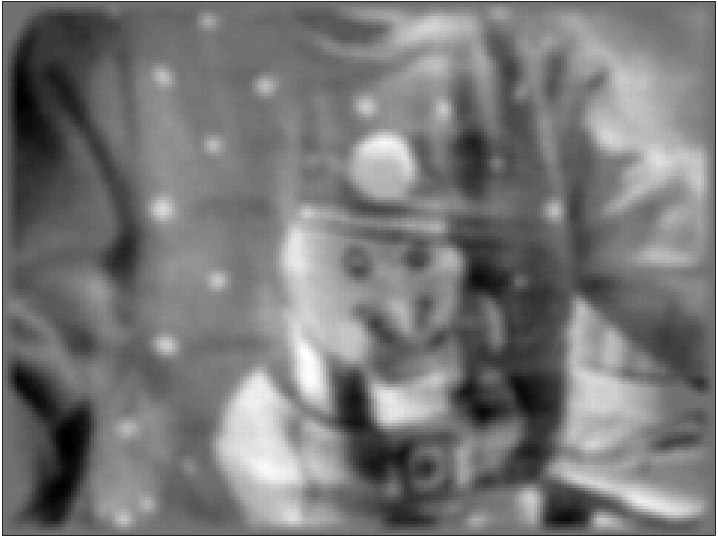}
		&
		\includegraphics[width=\kernelwidth]{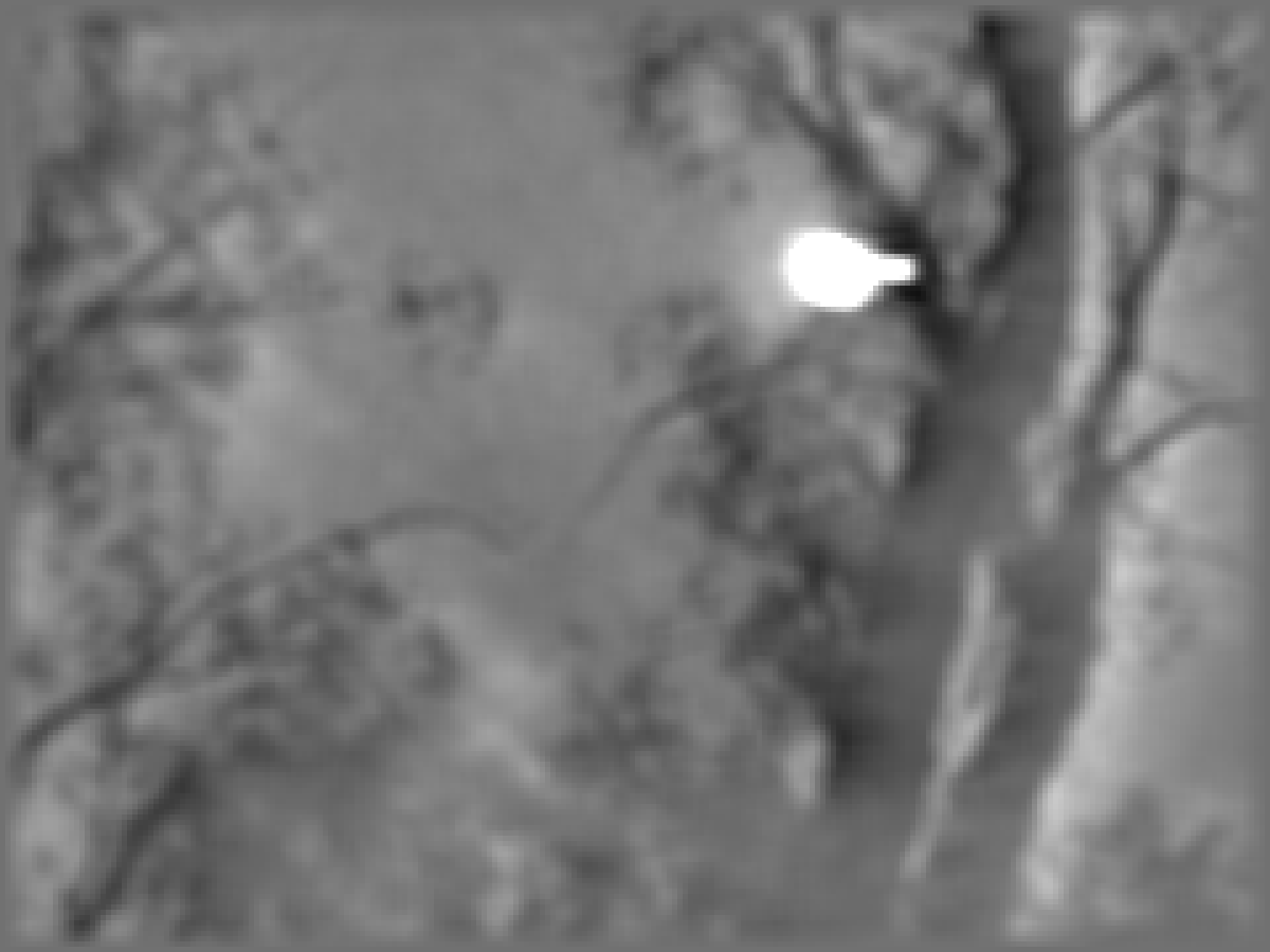}
		&
		\includegraphics[width=\kernelwidth]{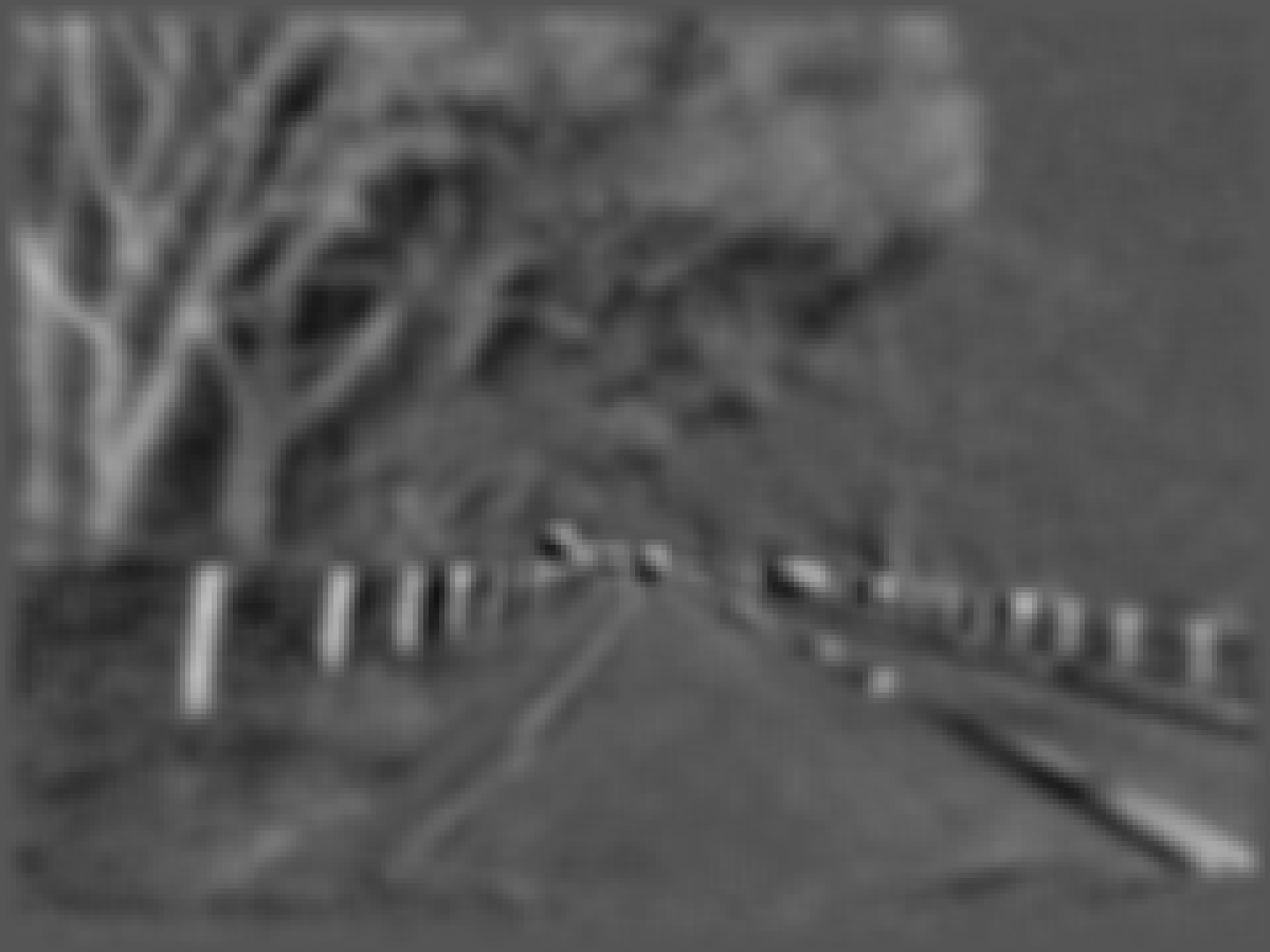}
		\\
		Sobel $x$
		
		\medskip
		$
		\begin{bmatrix}
		-1 & 0 & 1 \\
		-2 & 0 & 2 \\
		-1 & 0 & 1 
		\end{bmatrix}
		$ &
		\includegraphics[width=\kernelwidth]{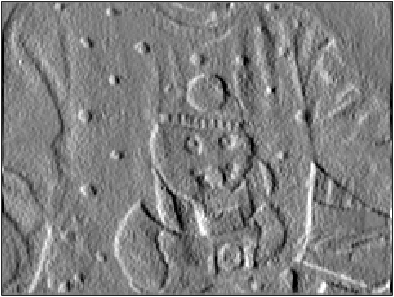}
		&
		\includegraphics[width=\kernelwidth]{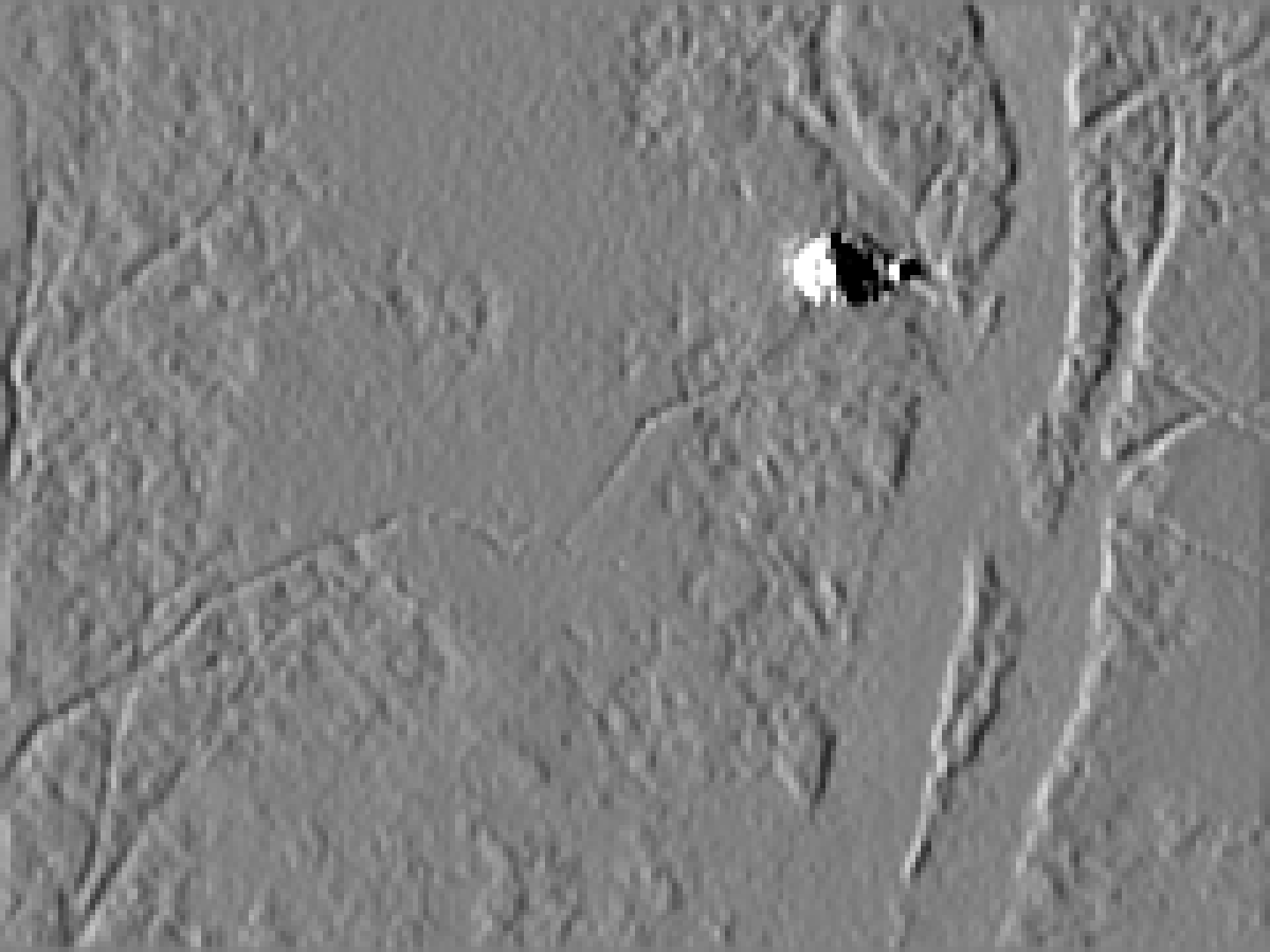}
		&
		\includegraphics[width=\kernelwidth]{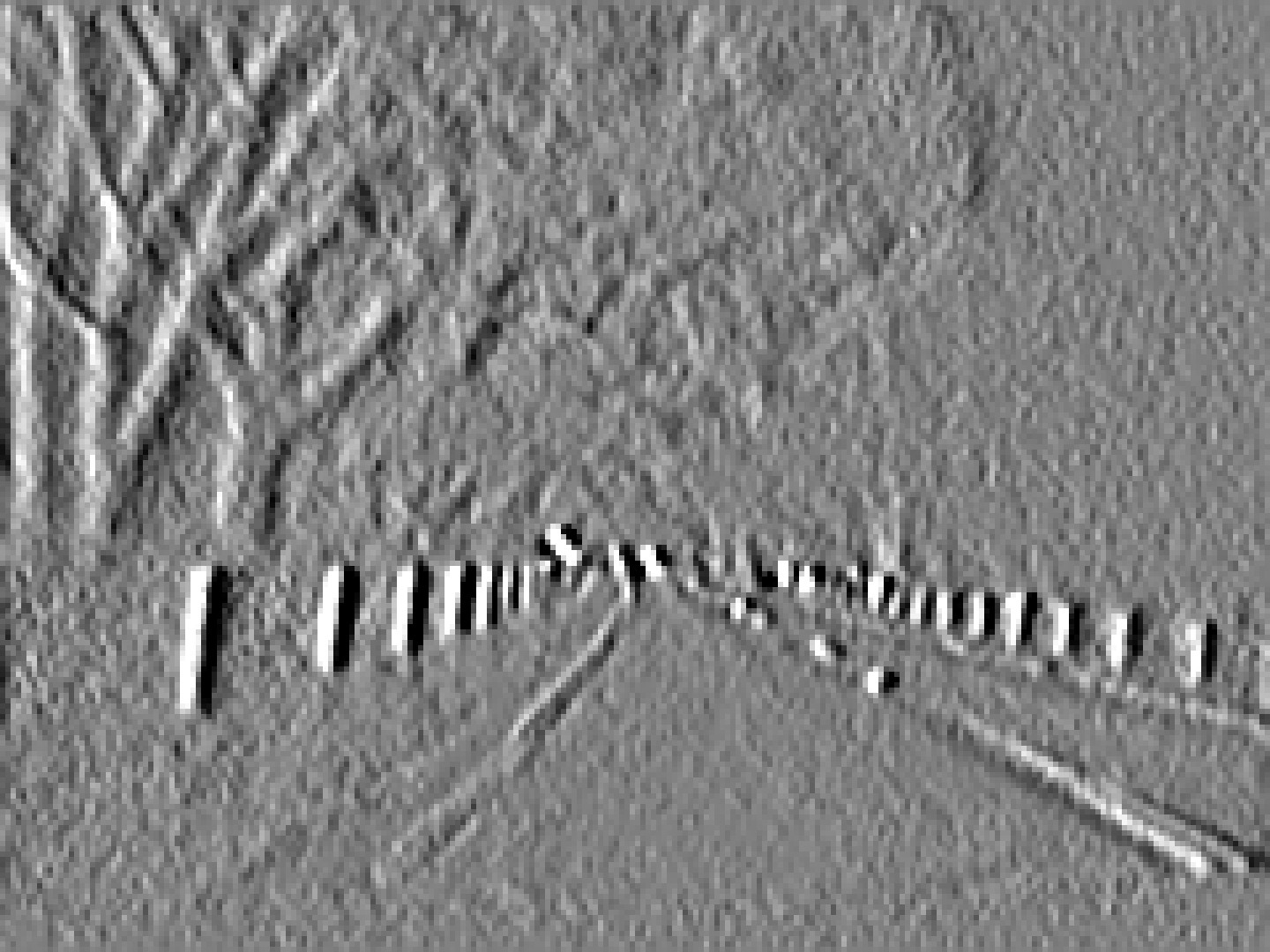}
		\\
		Sobel $y$
		
		\medskip		
		$
		\begin{bmatrix}
		-1 & -2 & -1 \\
		0 & 0 & 0 \\
		1 & 2 & 1 
		\end{bmatrix}
		$ &
		\includegraphics[width=\kernelwidth]{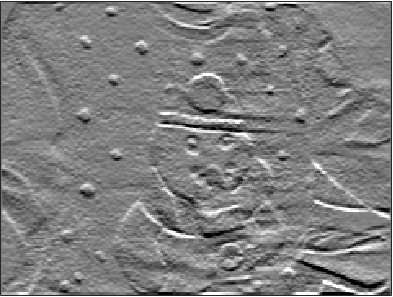}
		&
		\includegraphics[width=\kernelwidth]{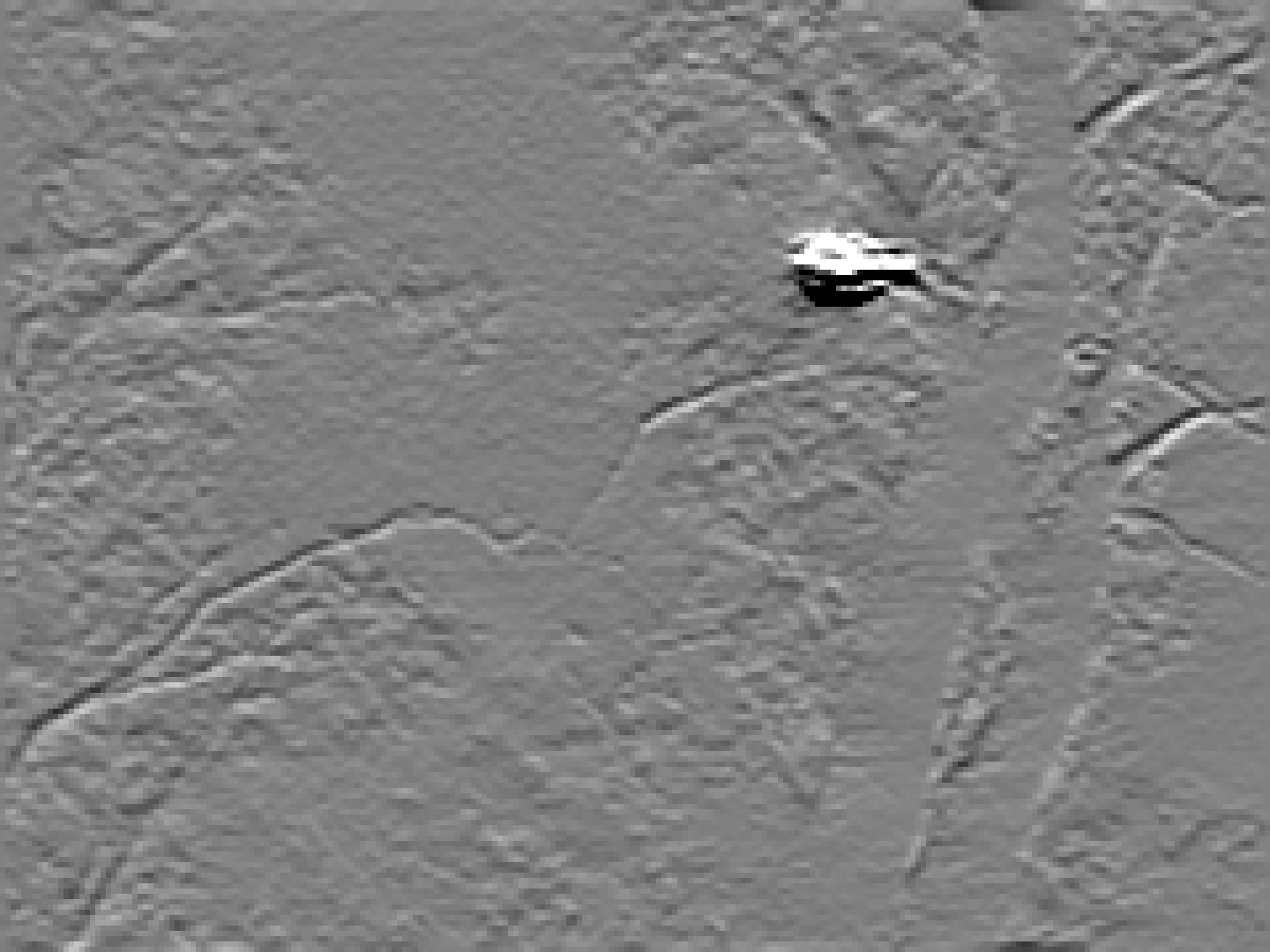}
		&
		\includegraphics[width=\kernelwidth]{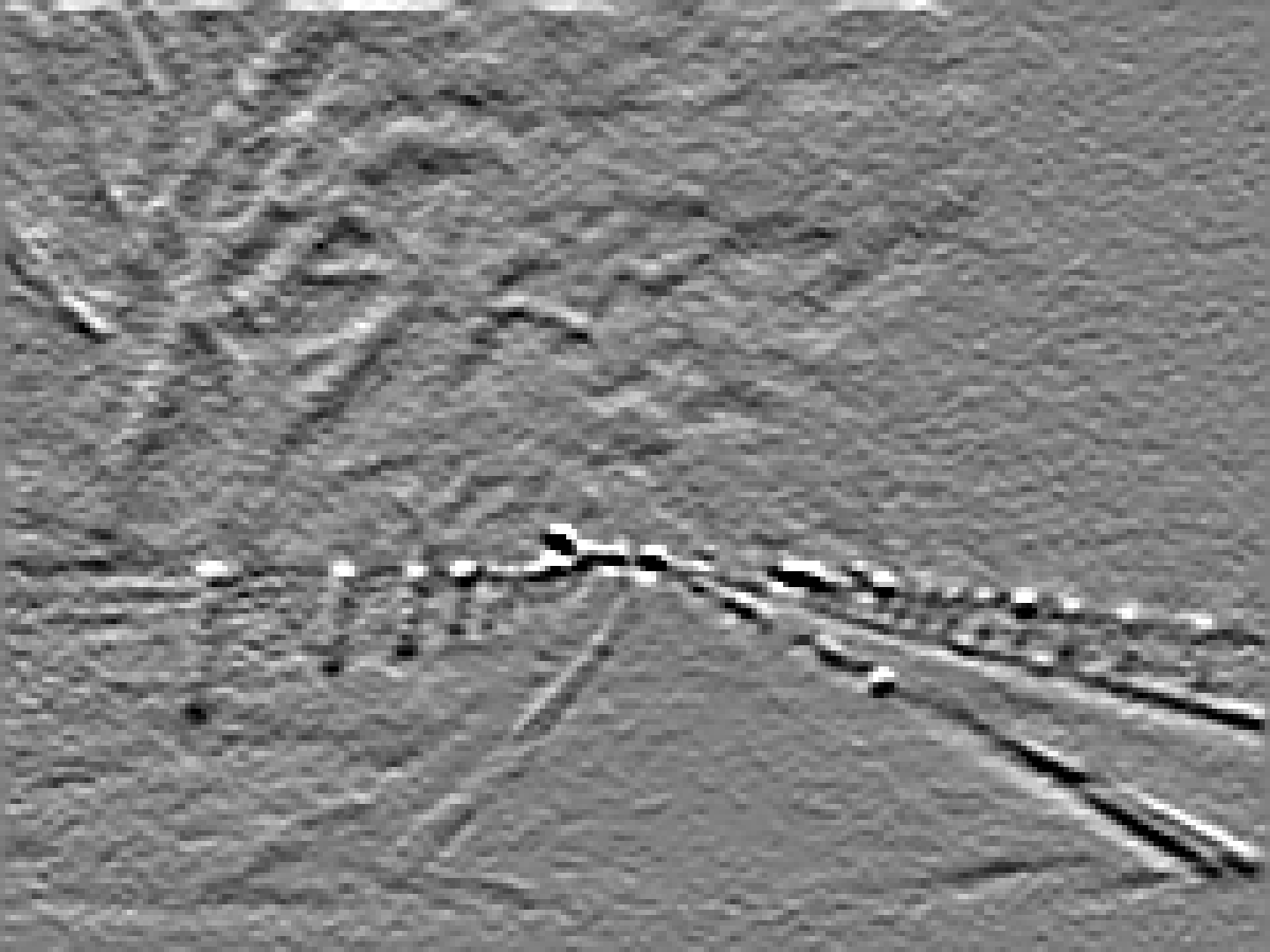}
		\\
		Laplacian
			
		\medskip
		$
		\begin{bmatrix}
		1 & 2 & 1 \\
		2 & -12 & 2 \\
		1 & 2 & 1 
		\end{bmatrix}
		$ &
		\includegraphics[width=\kernelwidth]{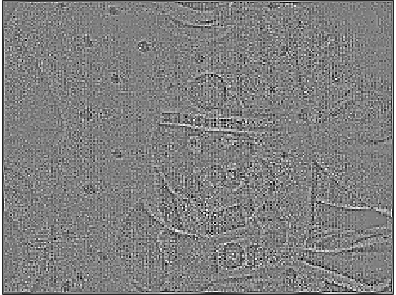}
		&
		\includegraphics[width=\kernelwidth]{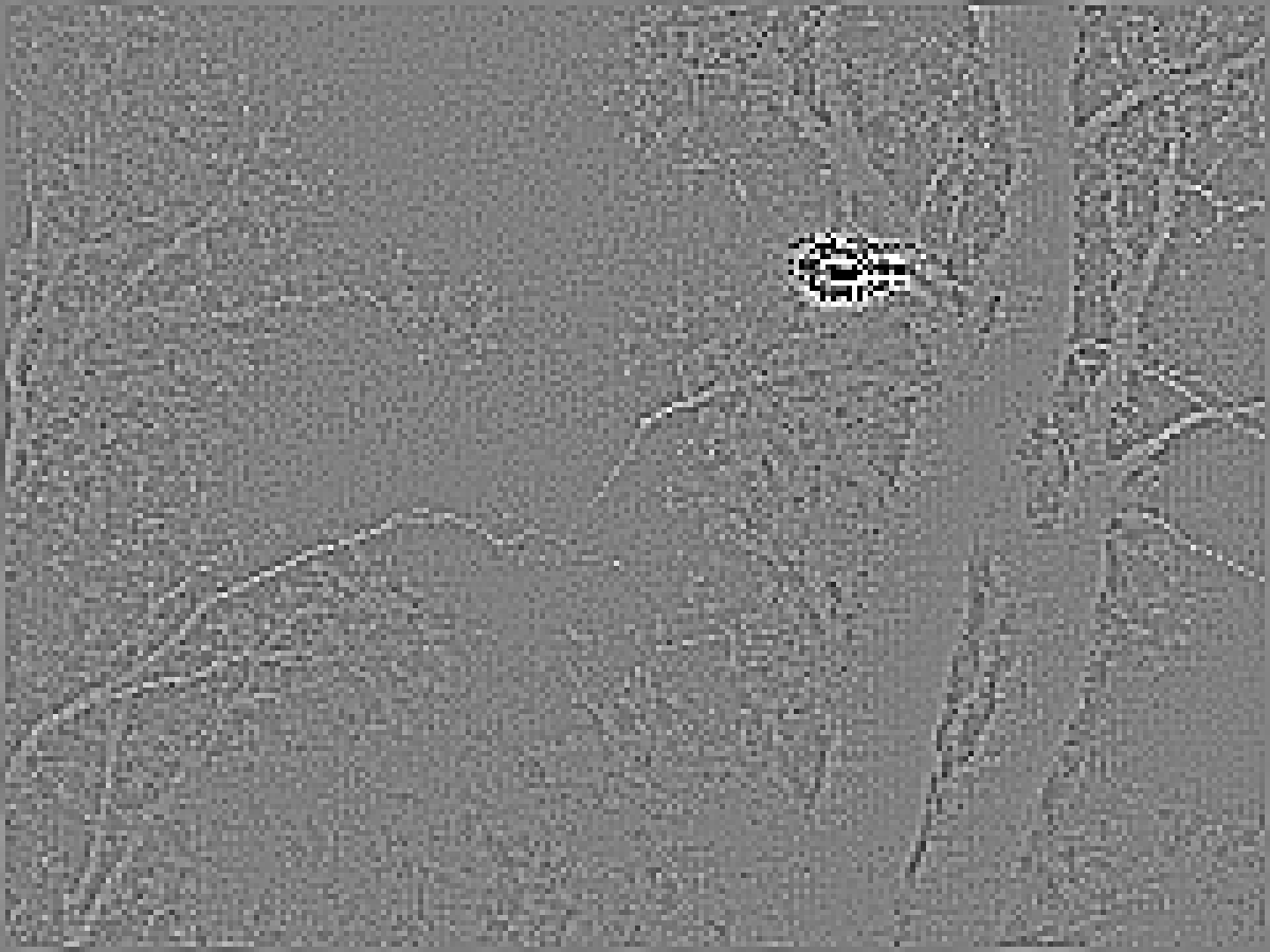}
		&
		\includegraphics[width=\kernelwidth]{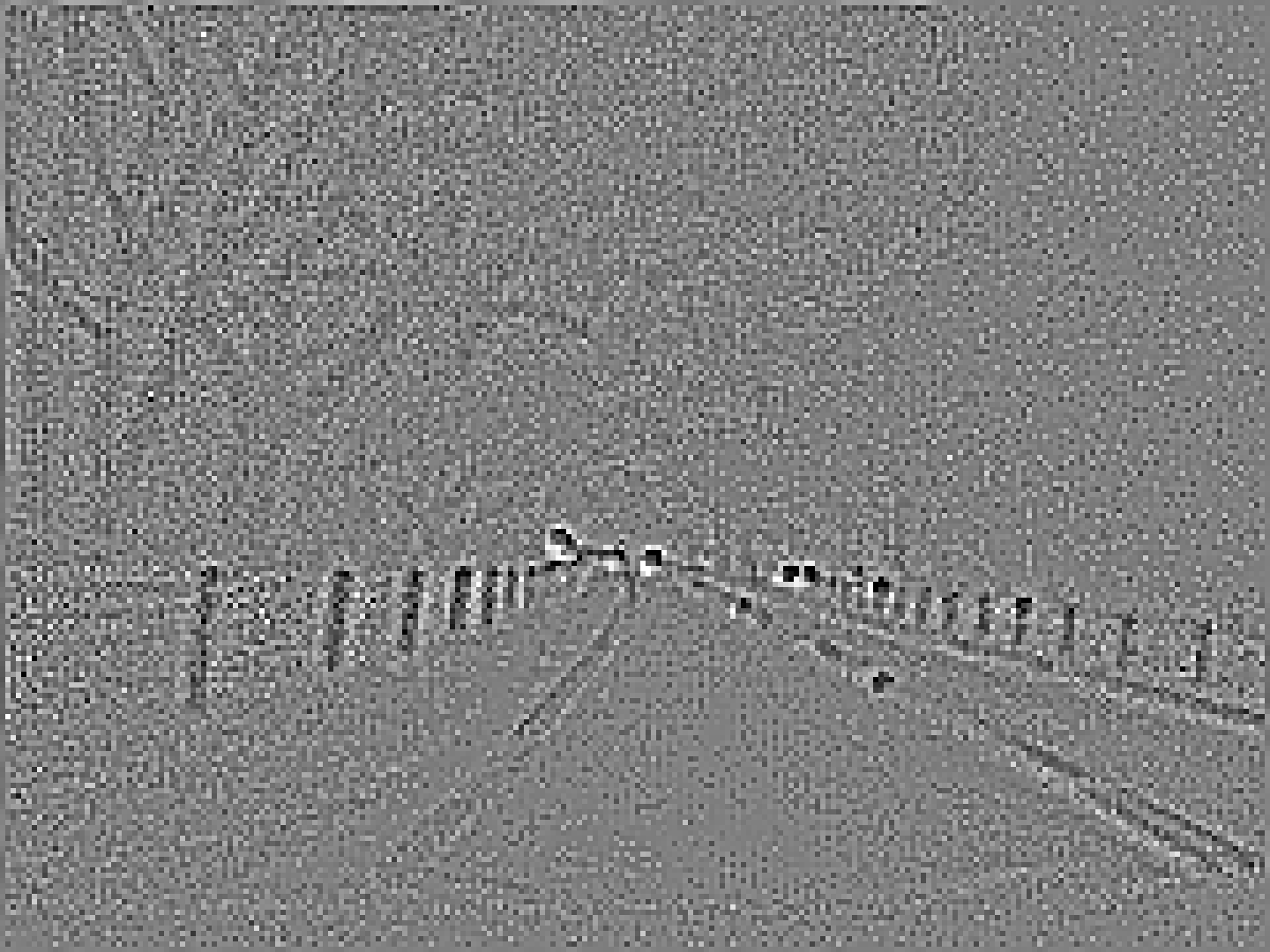}
		\\
		Poisson Reconstruction from Laplacian
		
		&
		\includegraphics[width=\kernelwidth]{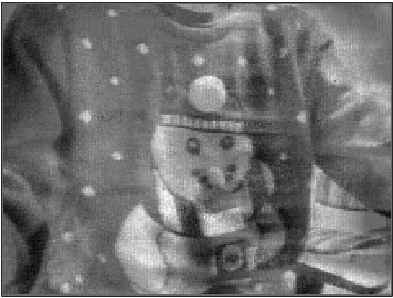}	
		&
		\includegraphics[width=\kernelwidth]{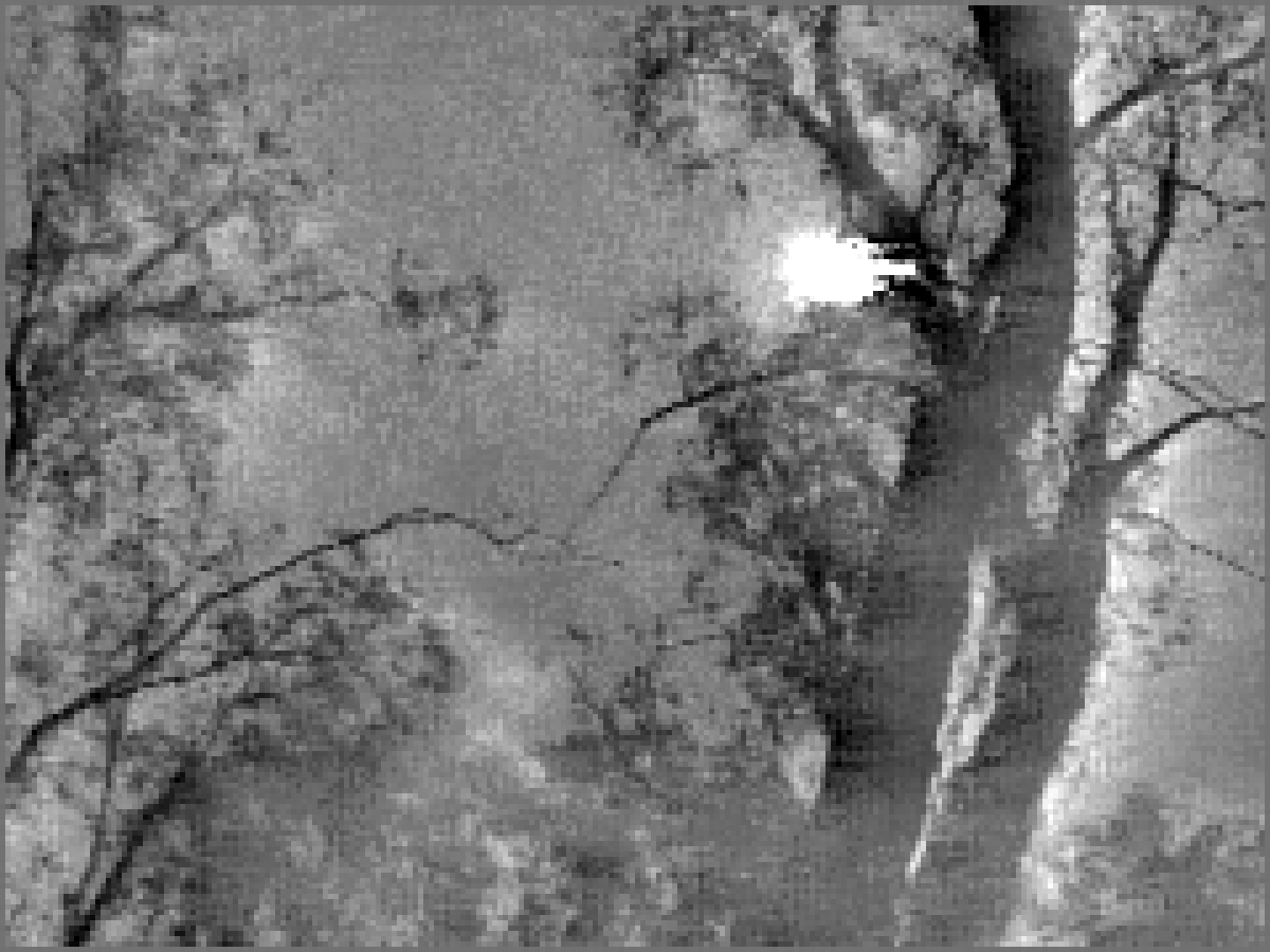}	
		&
		\includegraphics[width=\kernelwidth]{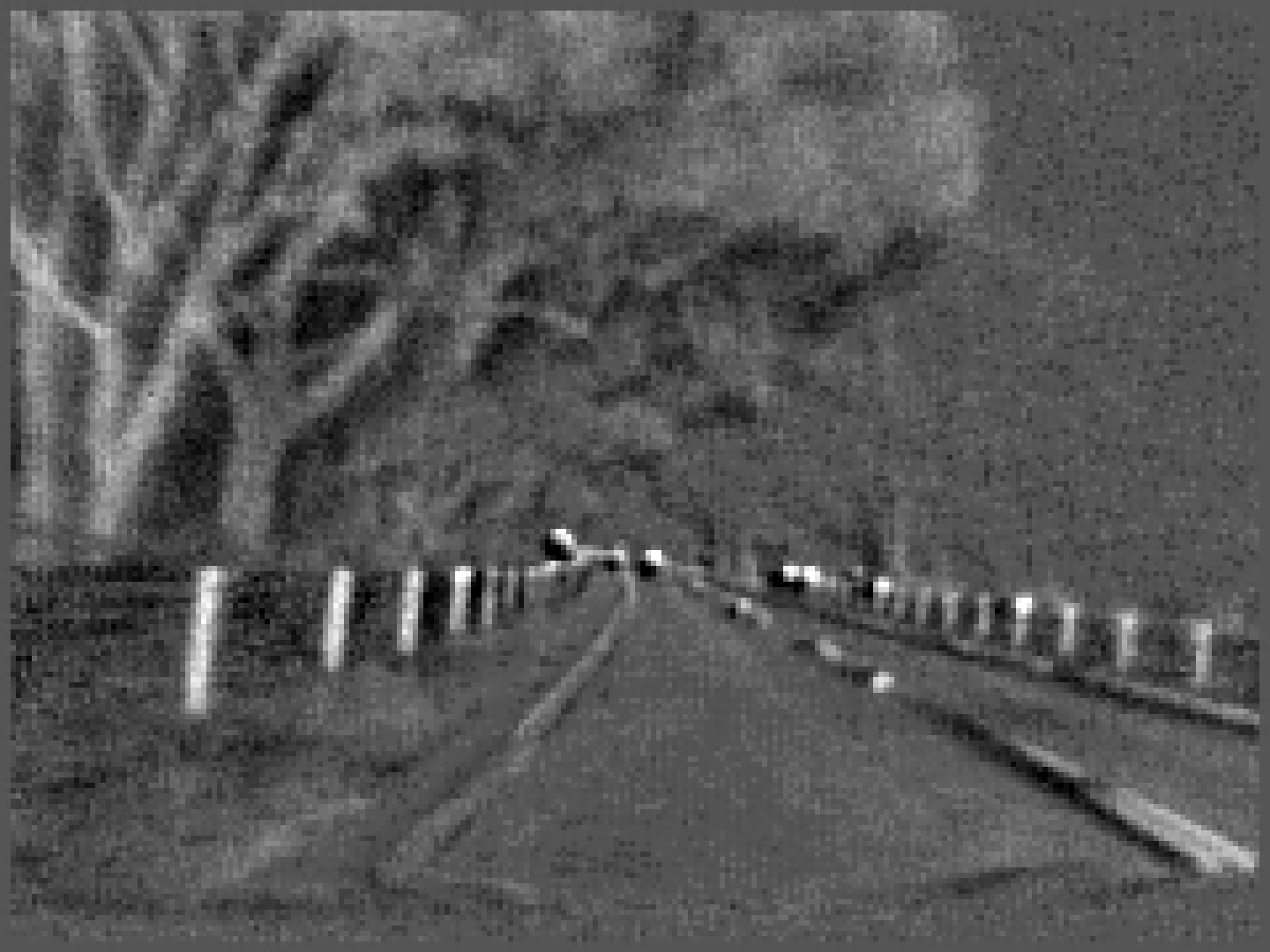}		
	\end{tabular}
}
\caption{Different kernels $K$ applied to events using high-pass filter \eqref{eq:high-pass}. \texttt{Sun} demonstrates robustness in extreme dynamic range scenarios and \texttt{night\_drive} is captured in pitch black conditions, demonstrating excellent performance in low-light settings thanks to the event camera.
}
\label{fig:kernels}
\end{figure*}
Figure \ref{fig:kernels} displays a range of different filtered versions of an input sequence (\texttt{sun} and \texttt{night\_drive} are taken from \cite{Scheerlinck18accv}).
The first row of Figure \ref{fig:kernels} shows the application of the identity kernel.
This kernel returns a (temporal) high-pass filtered version of the original image.
The sequences that follow, for a range of different kernels, are generated directly from events using the proposed algorithm and convincingly appear as one would expect if the kernel had been applied to the image reconstruction from the top row.
The key advantage of the proposed approach is that is does not incur latency or additional computation associated with reconstruction.
The sequences in Fig \ref{fig:kernels} are:
\begin{itemize}
\item \texttt{Snowman} \emph{(left)}:
The first author wearing a knitted jumper with prominent snowman and snowflakes design (taken under normal office conditions).
\item \texttt{Sun} \emph{(centre)}: Looking directly at the sun through the trees. Exemplifies high dynamic range performance of the camera.
\item \texttt{Night\_drive} \emph{(right)}: Country road at night with no street lights or ambient lighting, only headlights.
The car is travelling at 80km/h causing considerable motion in the scene. Exemplifies performance in high-speed, low-light conditions.
\end{itemize}

\noindent Despite noise in the event stream, our approach reproduces a high-fidelity representation of the scene.
It is particularly interesting to note the response for the two challenging sequences \texttt{sun} and \texttt{night\_drive}.
In both cases the image is clear and full of detail, despite the high dynamic range of the scene.

The second row computes a (spatial) low pass Gaussian filter of the sequences.
The low pass nature of the response is clear in the image.
The authors note that if it was desired to compute an image pyramid then it is a straightforward generalisation of the filter equations to reduce the state dimension at a particular level of the image pyramid by linear combination of pixel values.
The resulting filter would still be linear and the same filter equations would apply.

The third and fourth rows display the internal filter state for the Sobel kernels in both vertical and horizontal directions.
The results show that the derivative filter state is operating effectively even in very low light and high dynamic range conditions.

Rows five and six display the Laplacian of the image (the sum of second derivatives of the image) and a Poisson reconstruction built from the Laplacian image.
The Laplacian kernel computes an approximation of the divergence of the gradient vector field.
It can be used for edge detection: zero crossings in the Laplacian response correspond to inflections in the gradient and denote edge pixels.
It is also possible to reconstruct an original (log) intensity image from a Laplacian image using Poisson solvers \cite{Agrawal05iccv,Agrawal06eccv}.
In this case, we present the Poisson reconstruction of the Laplacian image (row six) primarily to verify the quality of the filter response.

It is important to recall that the internal state of the filter is computed directly from the event stream in all these cases.
For example, if only the Laplacian is required then there is no need to compute a grey scale image or gradient image.
\begin{figure*}[t]
	\centering
	\resizebox{0.9\textwidth}{!}{\begin{tabular}{ c c c c }
			Gradient & Corner State &\textbf{CHEC (ours)} & Harris \cite{Harris88}
			\\
			\includegraphics[width=\cornerwidth]{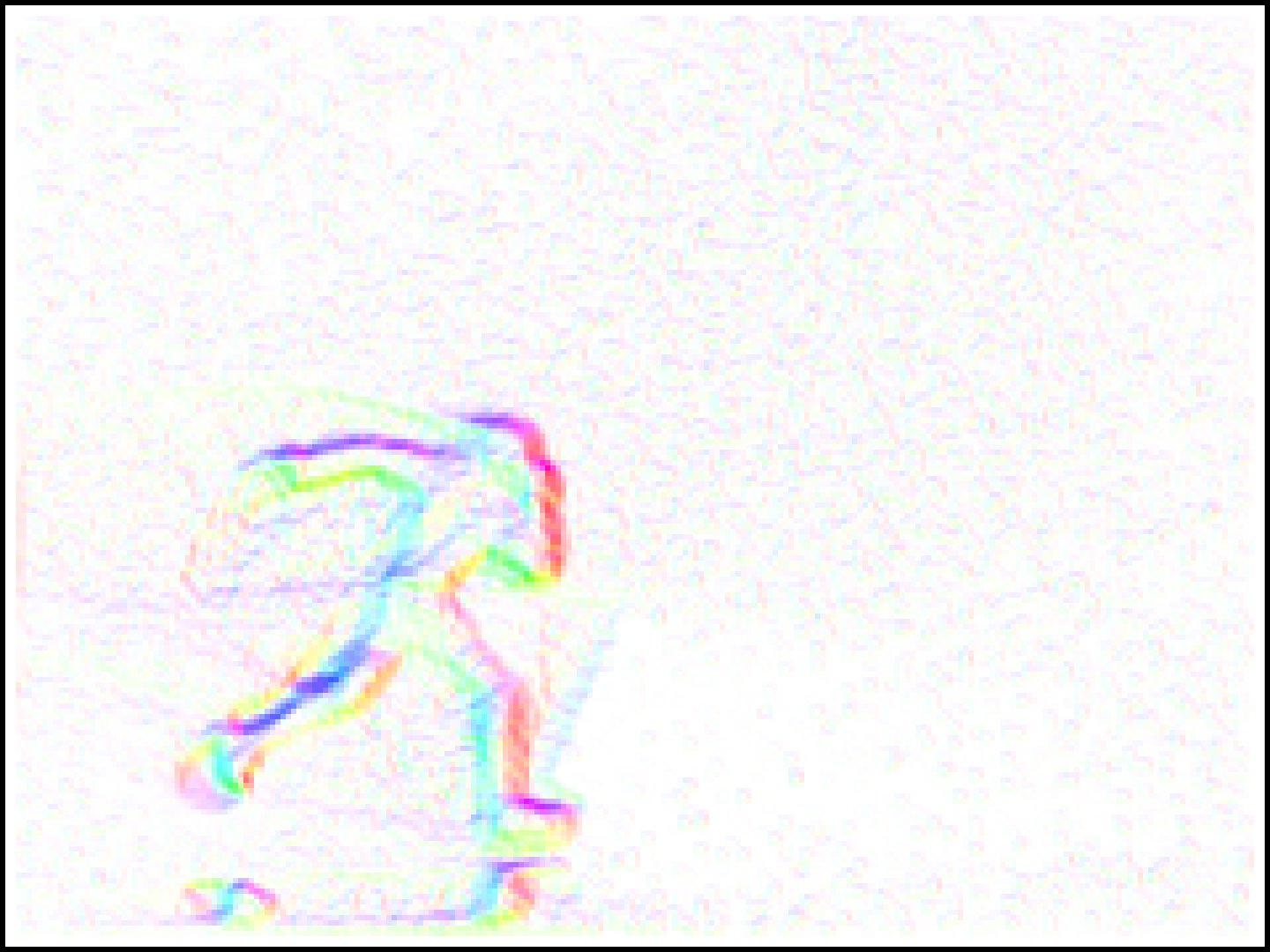}
			&
			\includegraphics[width=\cornerwidth]{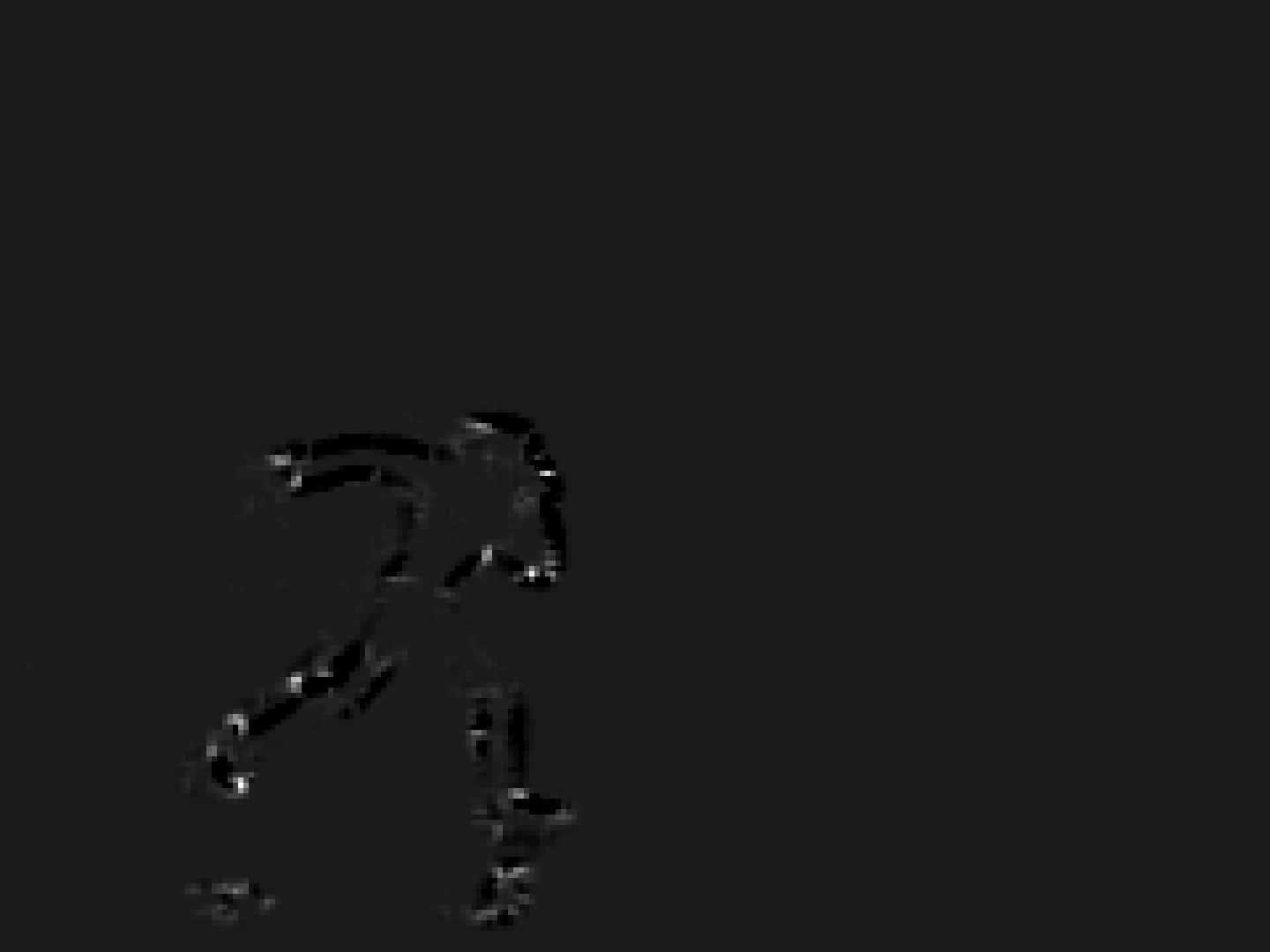}
			&
			\includegraphics[width=\cornerwidth]{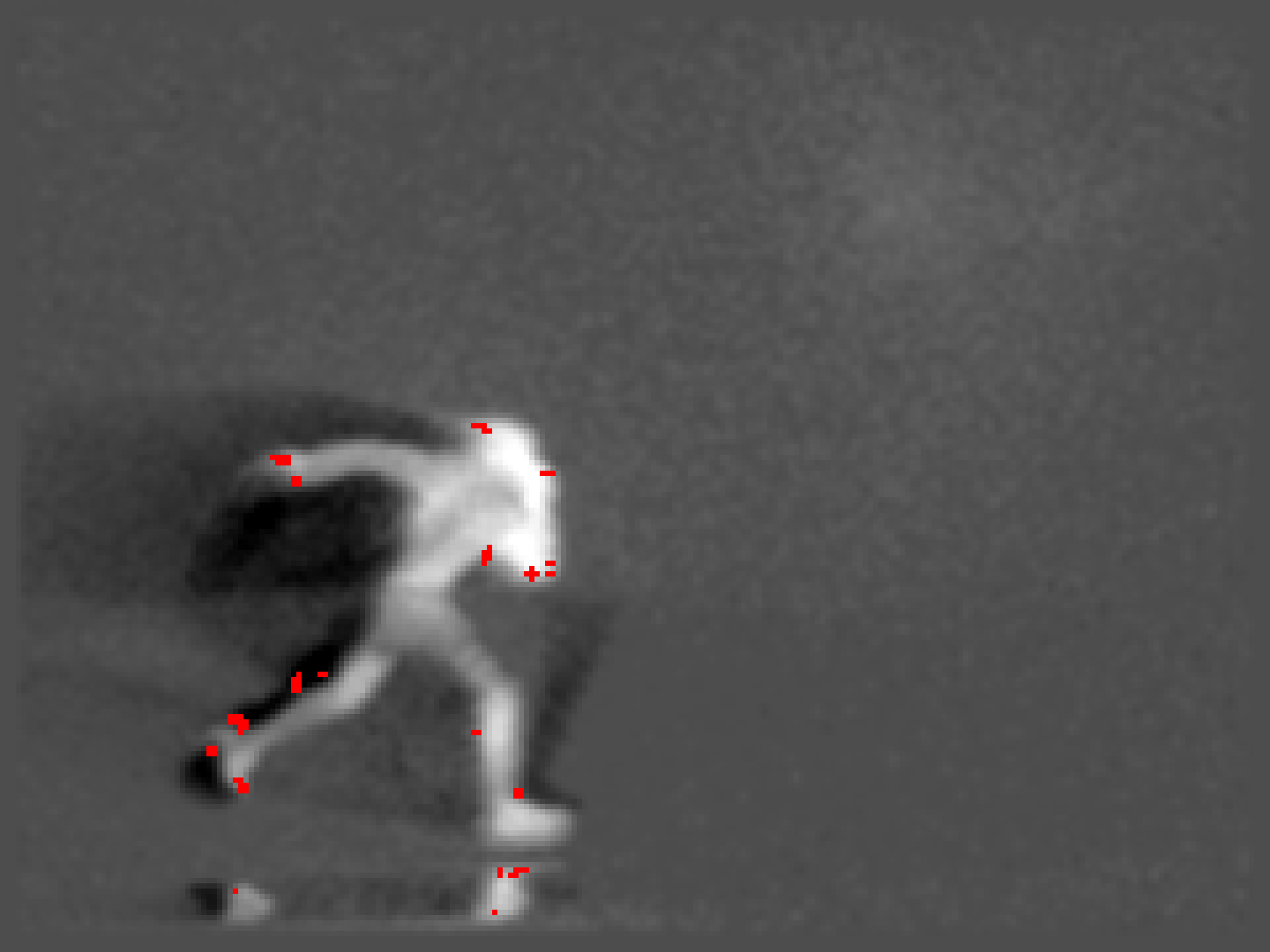}
			&
			\includegraphics[width=\cornerwidth]{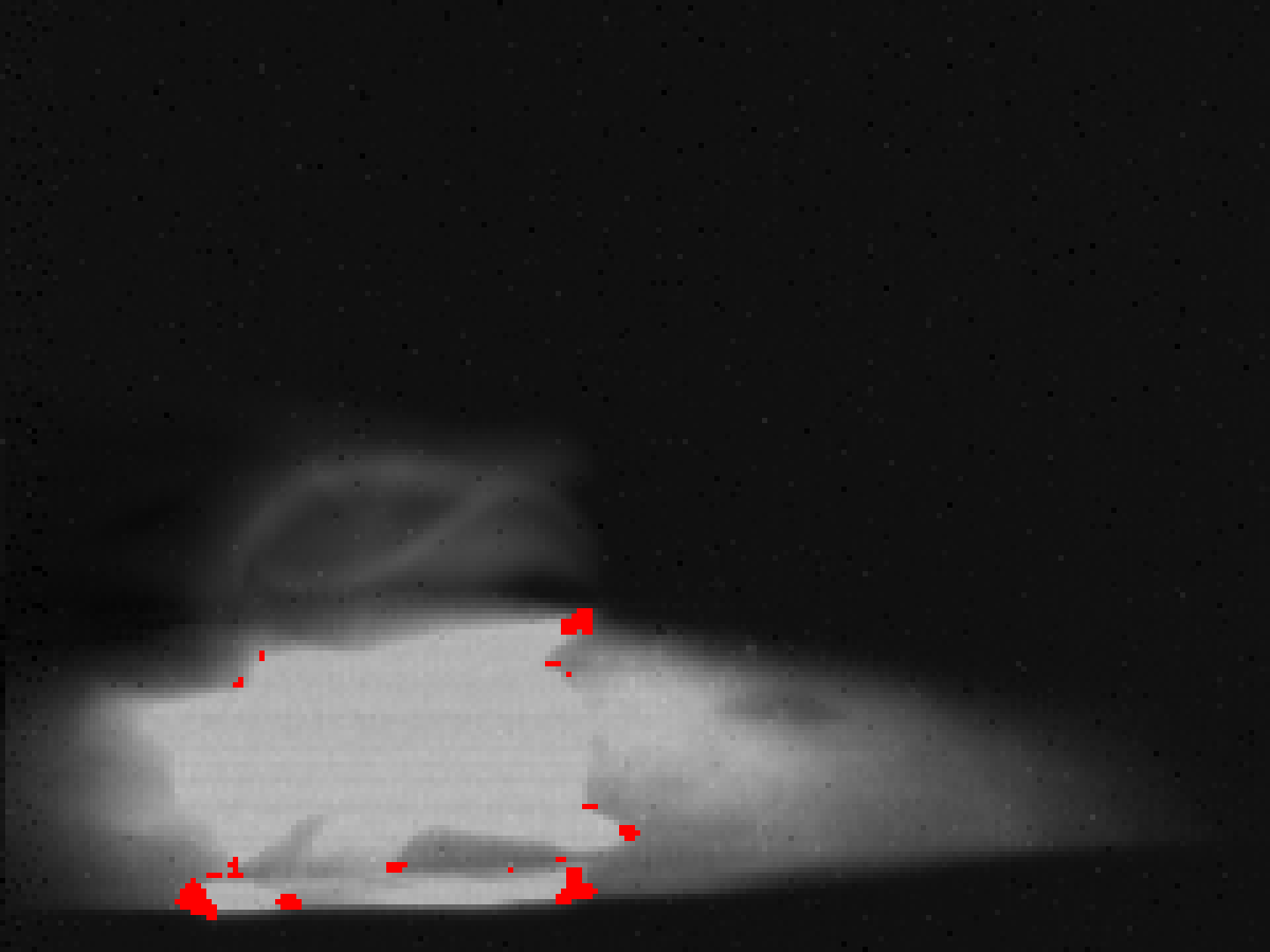}
			\\
			\includegraphics[width=\cornerwidth]{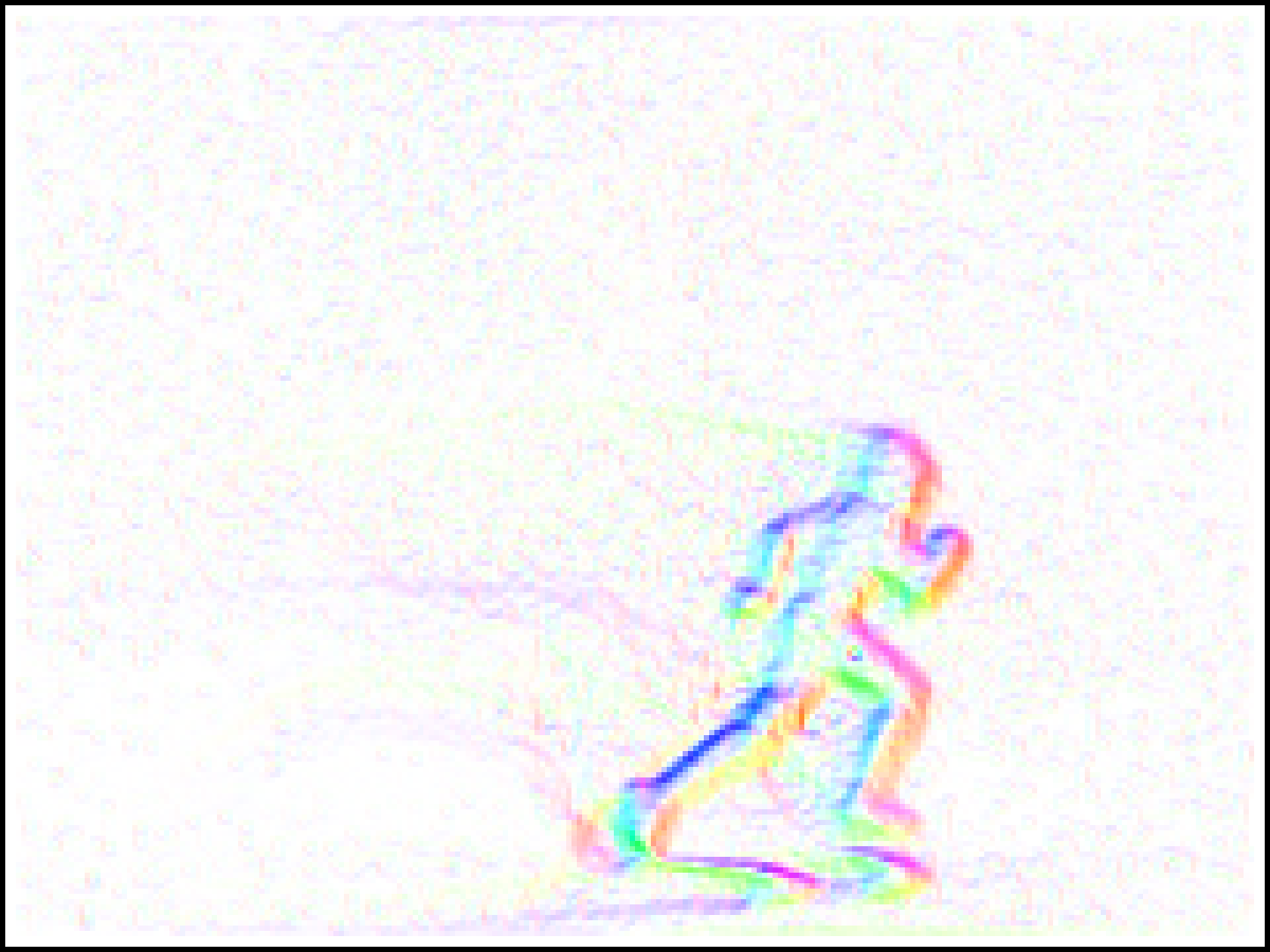}
			&
			\includegraphics[width=\cornerwidth]{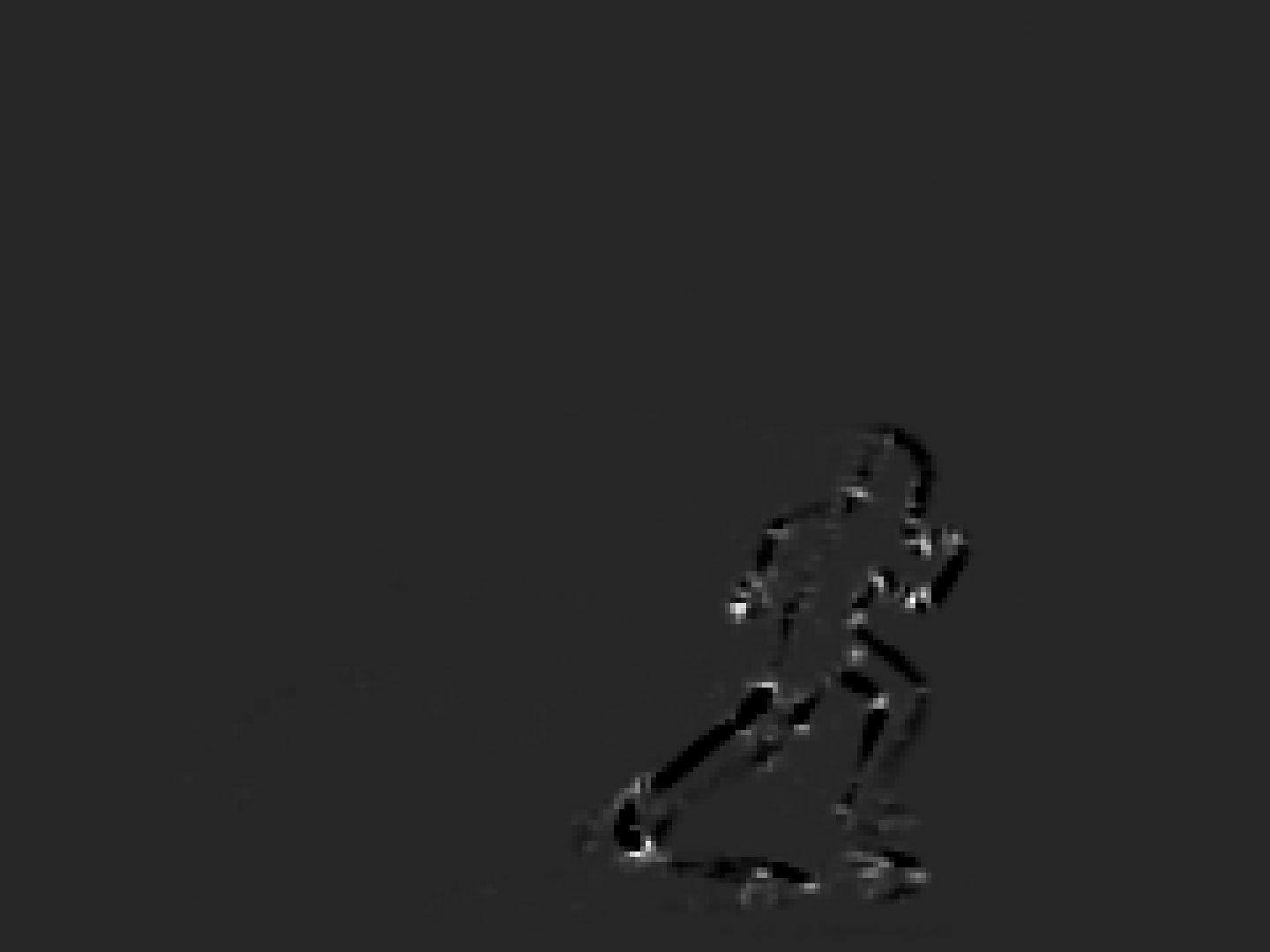}
			&
			\includegraphics[width=\cornerwidth]{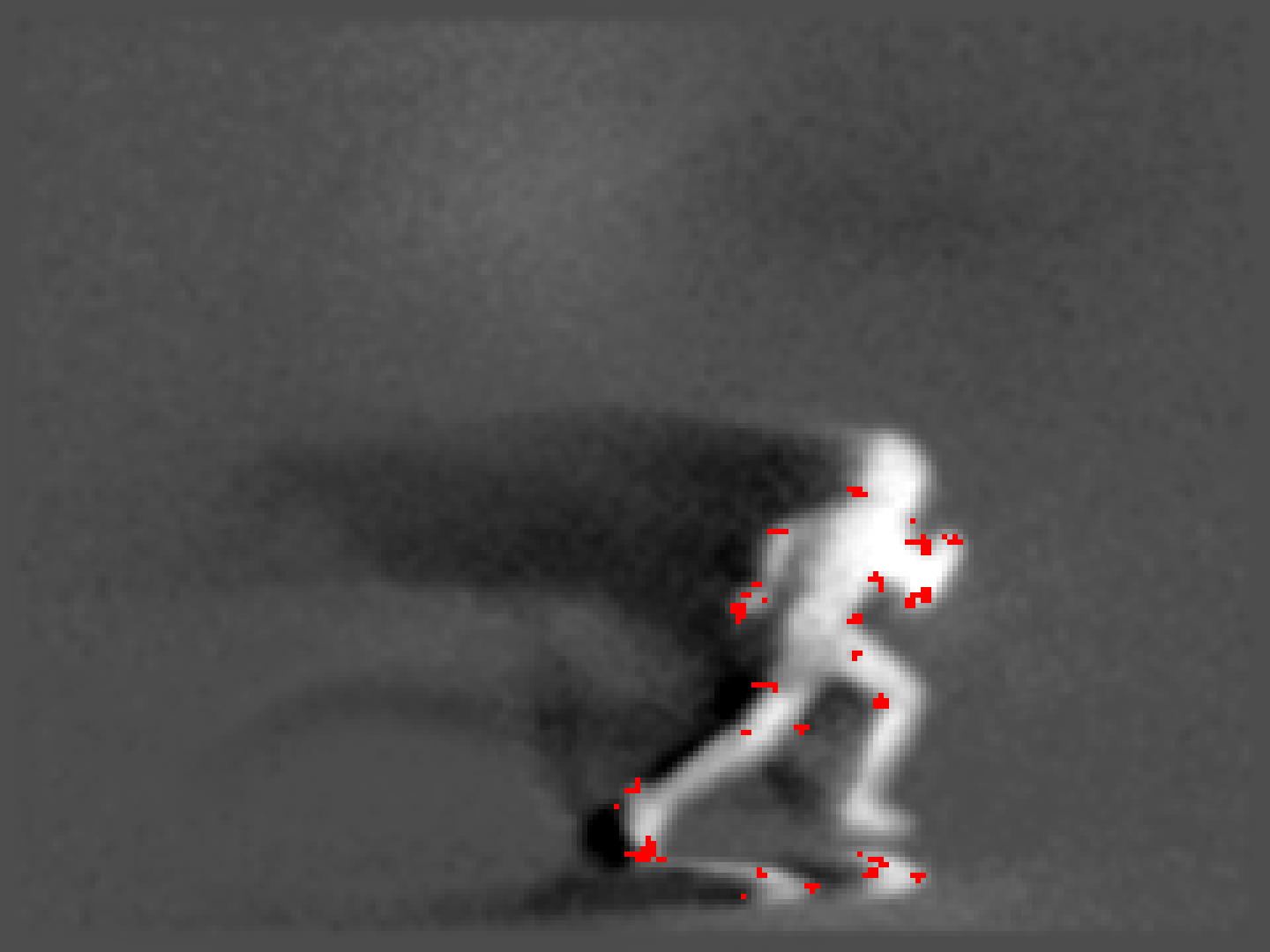}
			&
			\includegraphics[width=\cornerwidth]{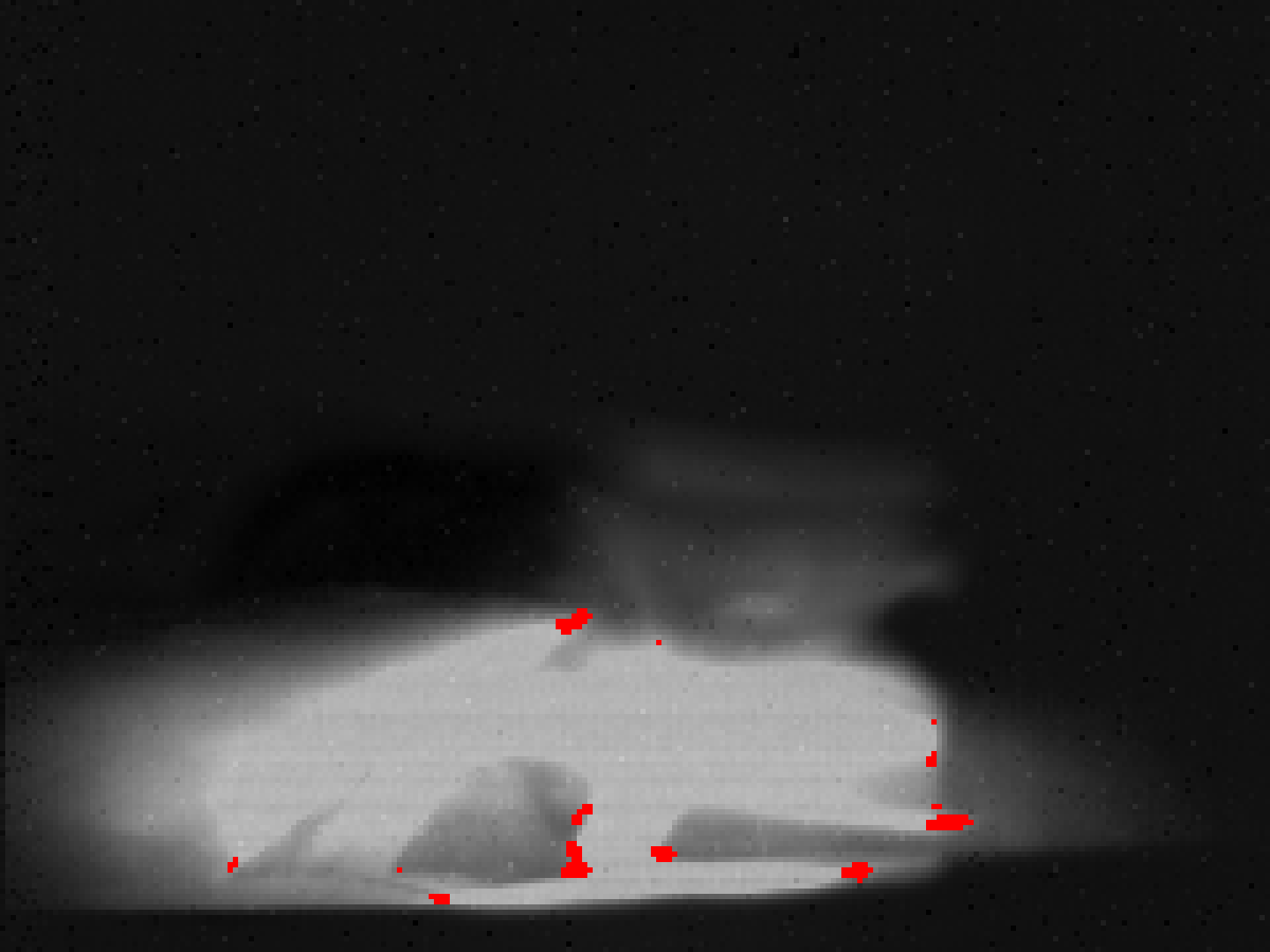}
			\\
			\includegraphics[width=\cornerwidth]{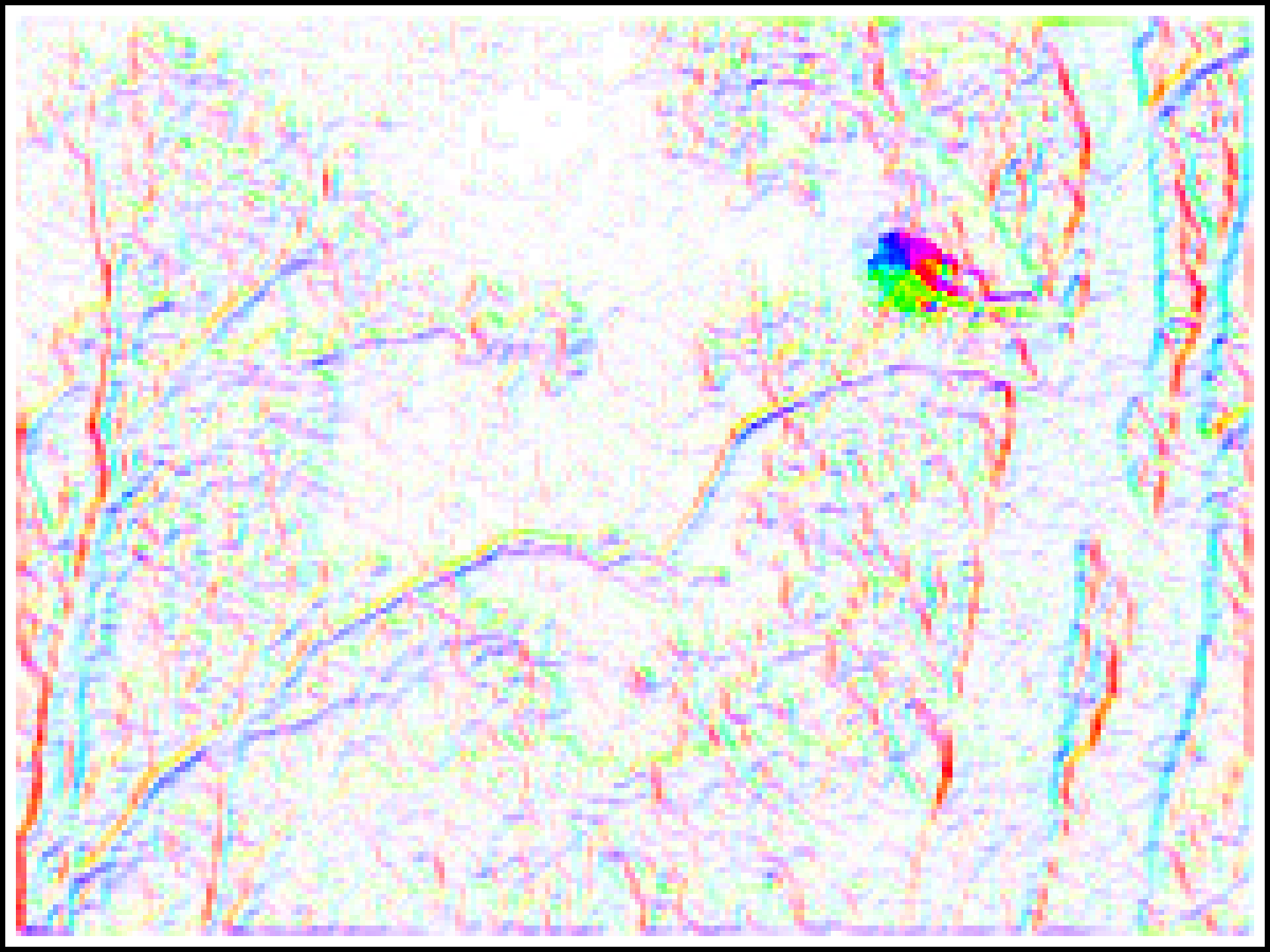}
			&
			\includegraphics[width=\cornerwidth]{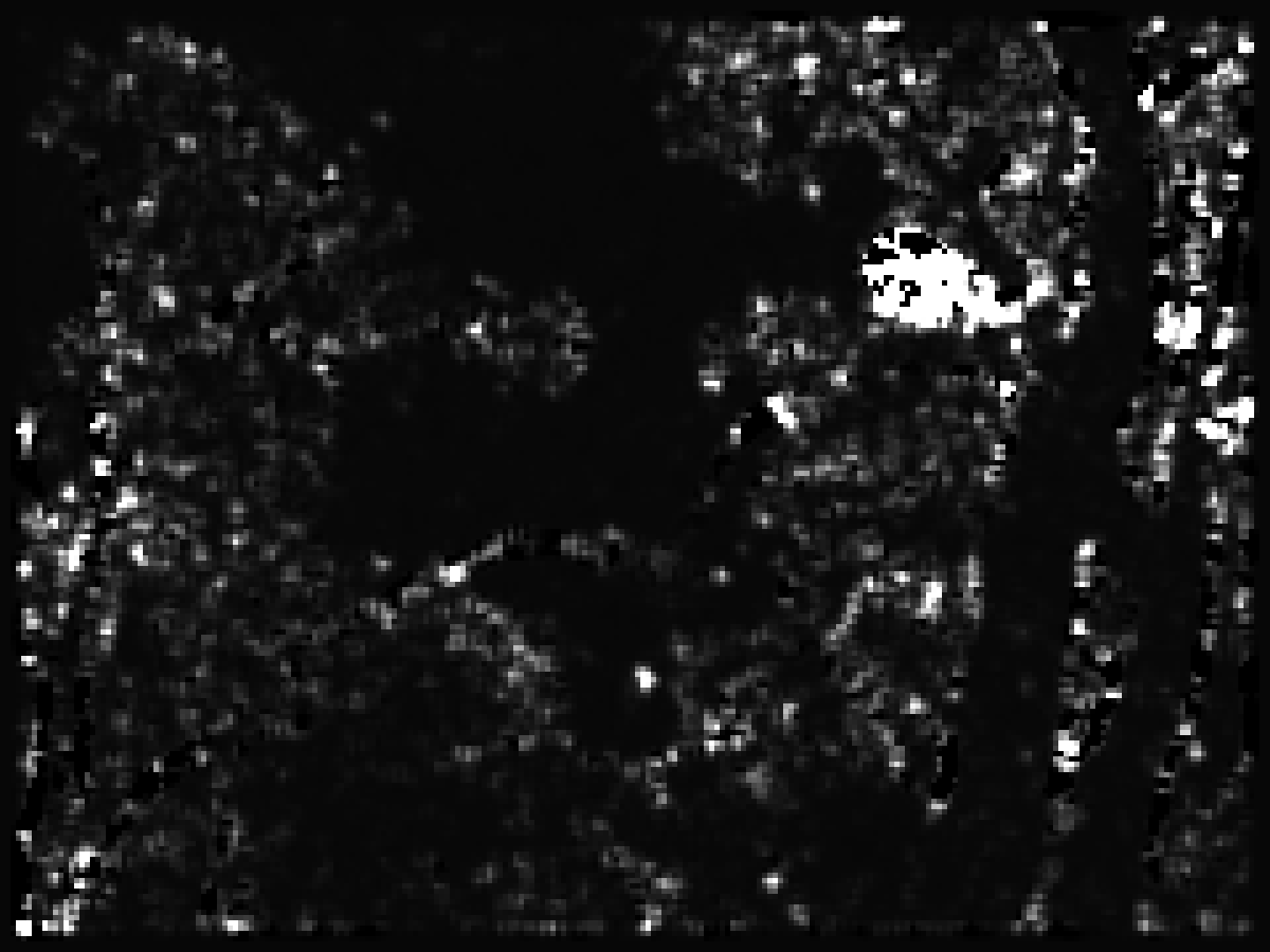}
			&
			\includegraphics[width=\cornerwidth]{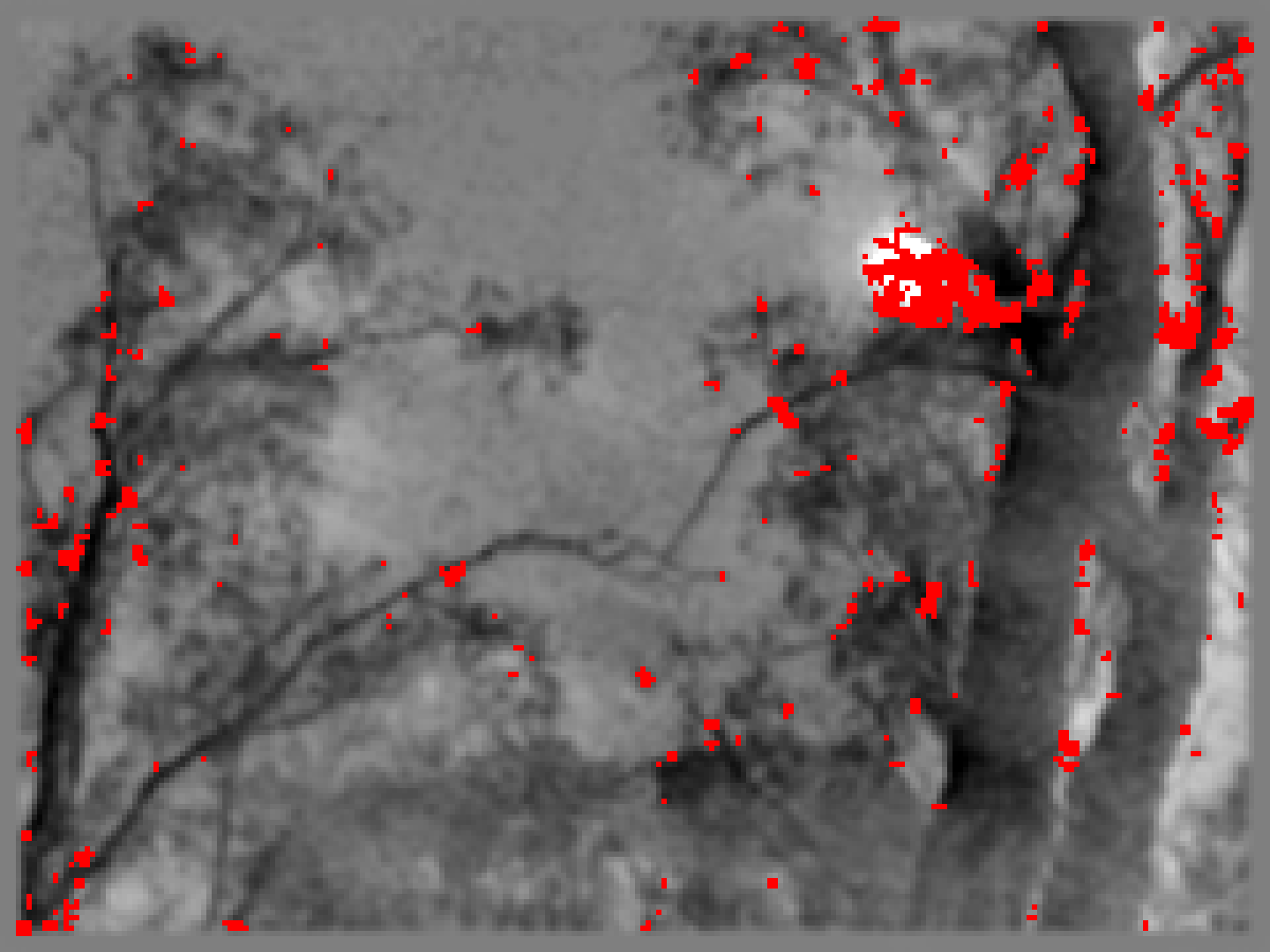}
			&
			\includegraphics[width=\cornerwidth]{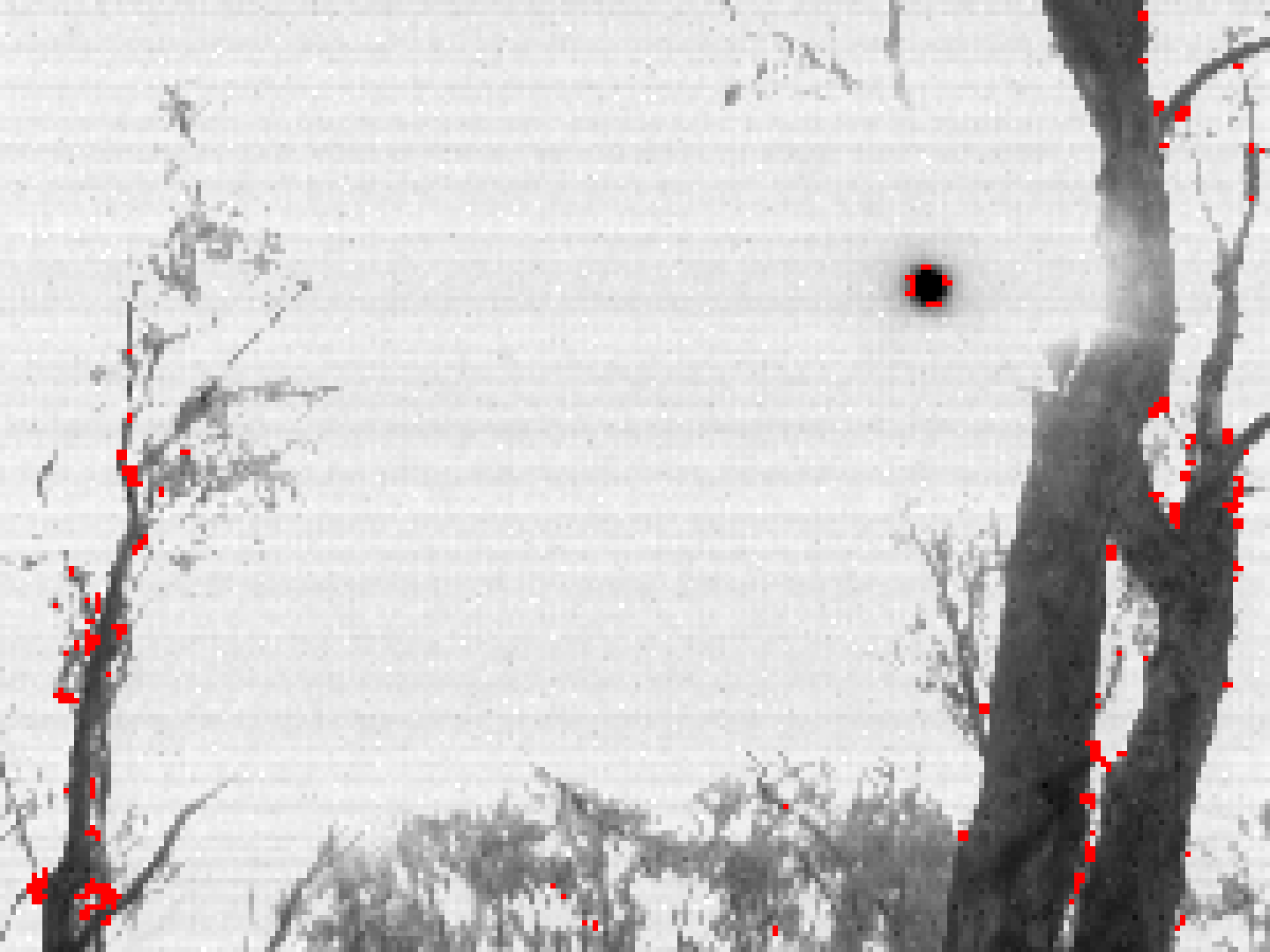}
			\\
			\includegraphics[width=\cornerwidth]{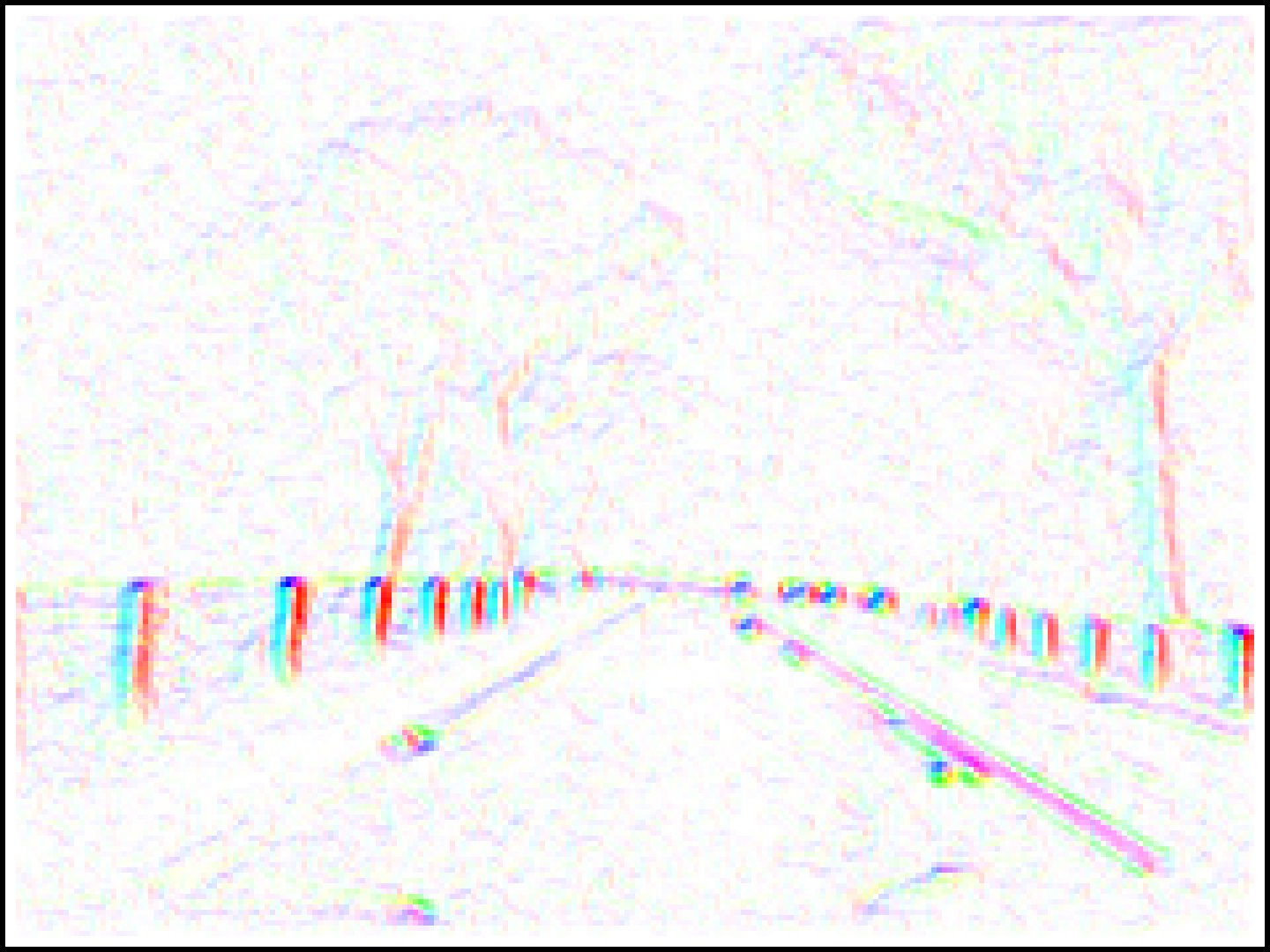}
			&
			\includegraphics[width=\cornerwidth]{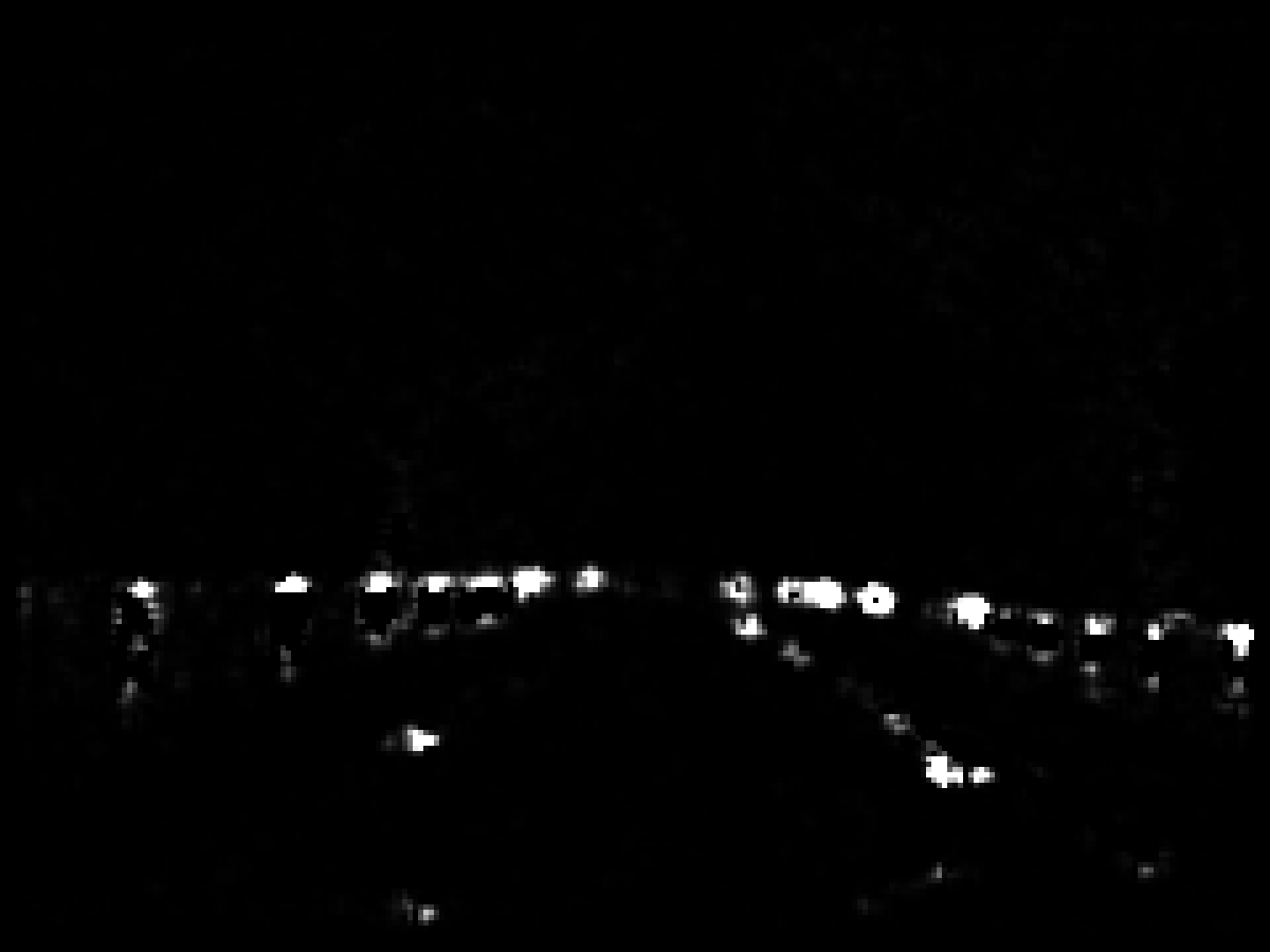}
			&
			\includegraphics[width=\cornerwidth]{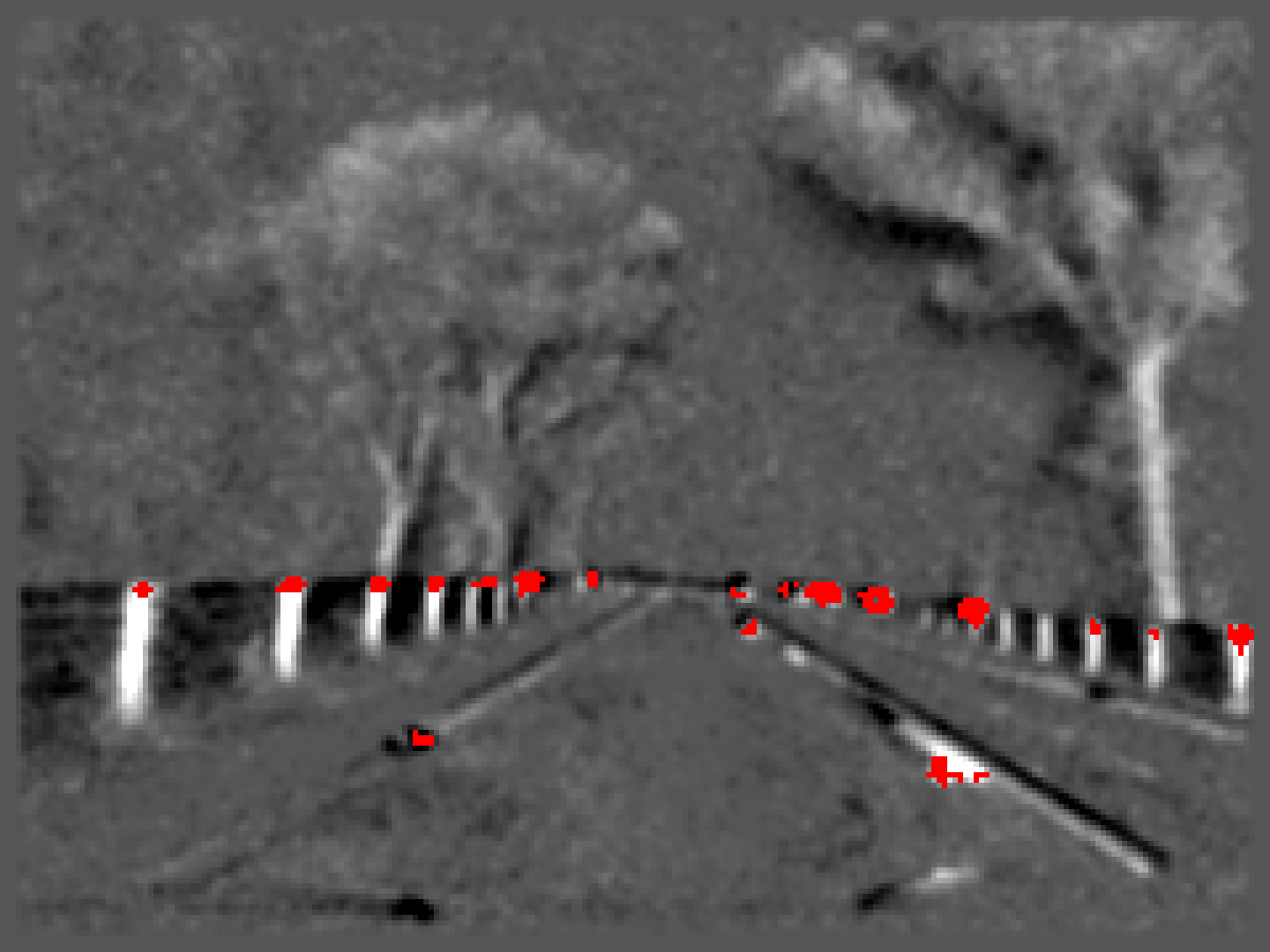}
			&
			\includegraphics[width=\cornerwidth]{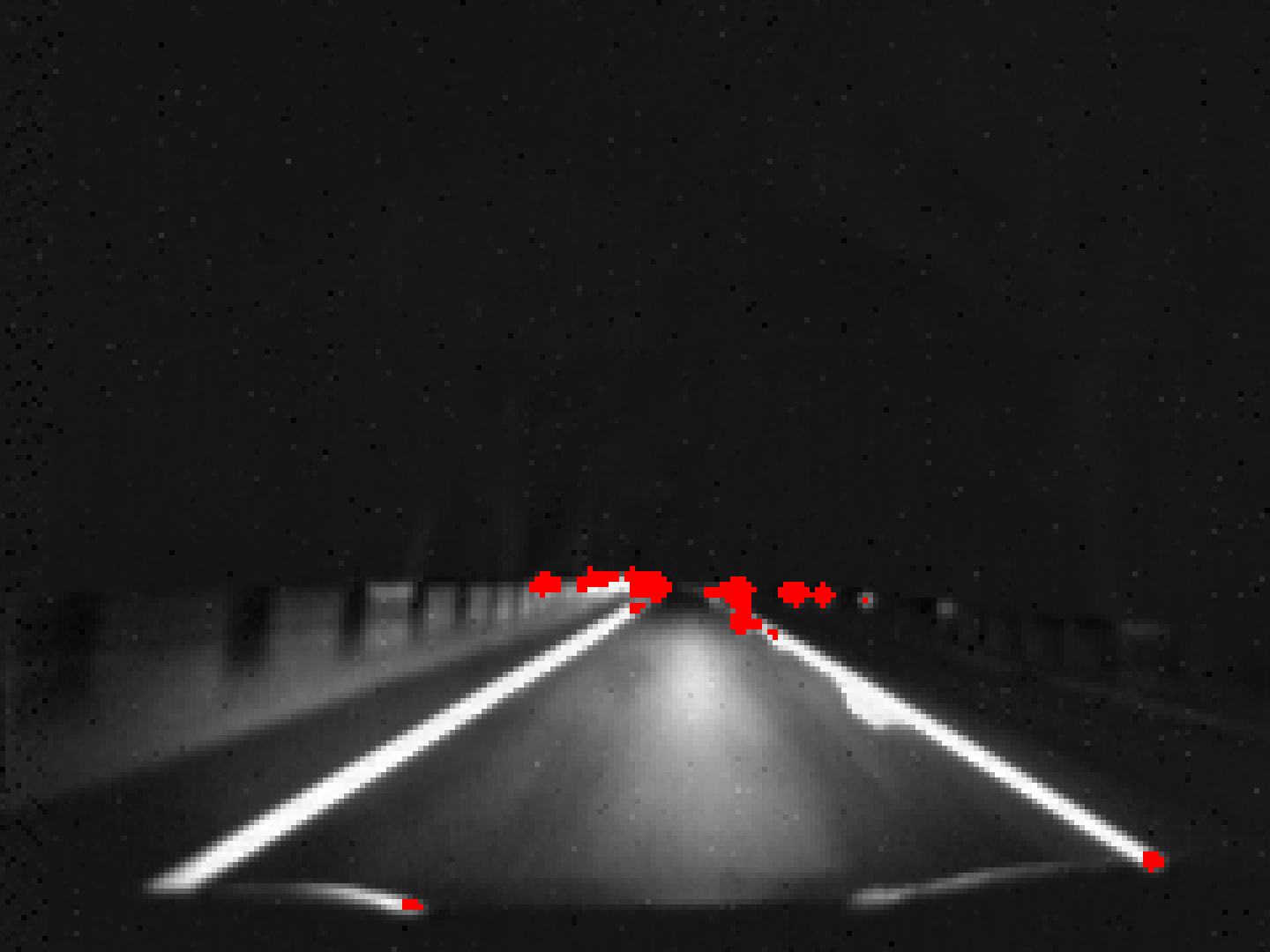}
		\end{tabular}	}
	\caption{Harris corner-response: Continuous Harris Event Corners (CHEC) applied to events verses conventional Harris applied to raw camera frames.
		Gradient shows a snap shot of the internal gradient state, obtained by applying Sobel $x$ and $y$ kernels directly to events.
		Corner State shows a snap shot of the full Harris corner-response state \eqref{eq:response} computed from only the gradient state.
		CHEC ({ours}) shows the corner-response thresholded at a suitable value and superimposed onto a log intensity image, obtained via Poisson reconstruction \cite{Agrawal05iccv,Agrawal06eccv} of Gradient.
		Note: log intensity is displayed purely for visualisation and is not used to compute corners.
		Harris shows the Harris detector \cite{Harris88} applied to raw image frames for comparison.
	}
	\label{fig:corner}
\end{figure*}
\subsection{Continuous Harris Event Corners (CHEC)}
\label{sec:corner}

To demonstrate a practical application of our filter we consider applying the image processing framework to detection of Harris corners.
We compute a continuous-time 2D state asynchronously that encodes the Harris-corner-response \cite{Harris88} of pixels.
Image gradients are computed using the filter architecture proposed in
equations \eqref{eq:ccode}, \eqref{eq:add_event} and \eqref{eq:timestamp} for Sobel kernels $K_x$ and $K_y$.
When a pixel gradient is updated then Harris corner response at that pixel location is also recomputed.
A threshold is applied at the pixel level and then non-maximum suppression is  applied locally to determine individual corners.
We call corners detected using this algorithm \emph{Continuous Harris Event Corners} (CHEC) since they are derived from a continuous-time Harris response image state.
A key advantage of our approach is that we are able to update the corner-response state asynchronously with each event, rather than having to synchronously update the entire state.

Let $\hat{G}(\bm{p}, t) = \big[ \hat{G_x}(\bm{p}, t) \quad \hat{G_y}(\bm{p}, t) \big]^\top$ denote an internal gradient state \eqref{eq:high-pass}.
The Harris matrix is
\begin{align}
M(\bm{p}, t) &:=
W * \hat{G}(\bm{p}, t) \hat{G}(\bm{p}, t)^T.
\end{align}
where $W$ is a smoothing kernel, e.g. box or Gaussian.
The Harris corner-response \cite{Harris88} is
\begin{align} \label{eq:response}
R(\bm{p}, t) &:= \det(M(\bm{p}, t)) - \gamma \text{ trace}(M(\bm{p}, t))^2,
\end{align}
where $\gamma$ is an empirically determined constant, in this case $\gamma = 0.04$.

Figure \ref{fig:corner} shows the continuous-time Harris corner-response state \eqref{eq:response} computed from the gradient state estimate \eqref{eq:high-pass}, on real sequences from \cite{Scheerlinck18accv}.
Column one shows the image gradient state. Column two shows the Harris corner-response state as a continuous-valued image state. Column three shows the binary output after applying a threshold to column two, overlaid on log intensity of the image (obtained via post processing of the stored gradient state and Poisson reconstruction) for visualisation purposes.
We emphasise that the image intensity was not required and not computed.
The final column shows the raw conventional frames and the binary output of the thresholded classical Harris response \cite{Harris88}.

\texttt{Night\_run} (top row Fig \ref{fig:corner}) is captured in pitch black conditions as someone runs in front of headlights of a car.
The conventional camera suffers extreme motion blur because of the high exposure-time required in low-light conditions.
Our approach leverages the high-dynamic-range, high-temporal-resolution event camera, yielding crisp edges and corners.
\texttt{Sun} (third row) displays artefacts in the corner state around the sun because of extreme brightness gradients caused by the high-dynamic range of the sun on a cloudless day, where the event camera is pushed to the limit.
Nevertheless, we still get clear corners around the branches and leaves of the trees, whereas the conventional camera frame is largely over-saturated.
\texttt{Night\_drive} (last row) demonstrates performance under challenging high-speed, low-light conditions.
Our approach clearly detects corners on roadside poles and road-markings.
The conventional camera frame is highly blurred and unable to detect corners in much of the image.
\begin{figure*}[t!]
	\centering
	\resizebox{\textwidth}{!}{\begin{tabular}{ c c c c c }
			\includegraphics[width=\comparewidth]{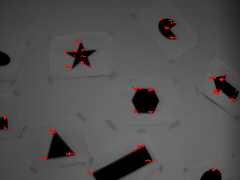}
			&
			\includegraphics[width=\comparewidth]{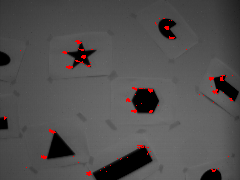}
			&
			\includegraphics[width=\comparewidth]{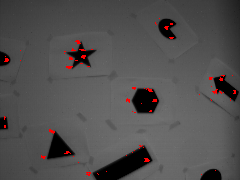}
			&
			\includegraphics[width=\comparewidth]{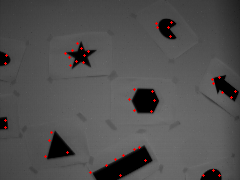}
			&
			\includegraphics[width=\comparewidth]{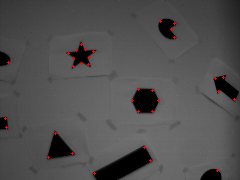}
			\\
			\includegraphics[width=\comparewidth]{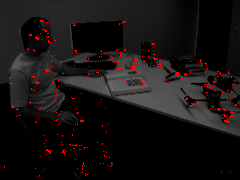}
			&
			\includegraphics[width=\comparewidth]{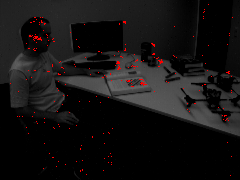}
			&
			\includegraphics[width=\comparewidth]{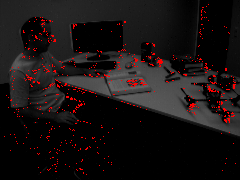}
			&
			\includegraphics[width=\comparewidth]{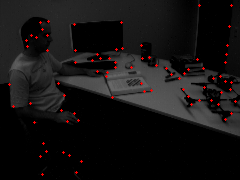}
			&
			\includegraphics[width=\comparewidth]{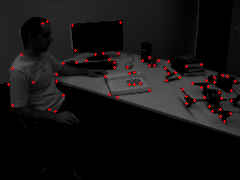}
			\\
			eHarris \cite{Mueggler17bmvc,Vasco16iros}
			&
			FAST \cite{Mueggler17bmvc}
			&
			ARC \cite{Alzugaray18ral}
			&
			\textbf{CHEC (ours)}
			&
			Harris \cite{Harris88}
		\end{tabular}	}
	\caption{Top: \texttt{shapes}, bottom: \texttt{dynamic}. We plot the raw camera frame for visualisation, but it is not used in any event-based corner detection (eHarris, FAST, ARC, CHEC). Corners appear shifted due to the low temporal resolution of the raw frame. Since eHarris, FAST and ARC provide asynchronous corner events, we accumulate the last 30ms for visualisation. Harris computes corners from the raw frames, which are subject to low temporal resolution and limited dynamic range.}
	\label{fig:comparison}
\end{figure*}

Figure \ref{fig:comparison} compares state-of-the-art event-based corner detectors \cite{Vasco16iros,Mueggler17bmvc,Alzugaray18ral}, as well as frame-based Harris detector (Harris) \cite{Harris88}, against our proposed method (CHEC), on real sequences from the event camera dataset \cite{Mueggler17ijrr} (\texttt{shapes\_translation} and \texttt{dynamic\_6dof}).
The corners identified are overlaid on the raw camera frame to improve visualisation of the results. 
eHarris was first developed by \cite{Vasco16iros}, and later improved by \cite{Mueggler17bmvc}.
We use improved eHarris code implementation of \cite{Mueggler17bmvc}.
We also compare against FAST event-based corner detector \cite{Mueggler17bmvc}  and ARC (asynchronous event-based corner detection) \cite{Alzugaray18ral}.
For state-of-the-art we use default parameters provided in the open-source code.
For CHEC, we increase the filter gain to $\alpha = 10 \,$rad/s to reduce low-temporal-frequency noise.
To extract corners from the Harris response of both Harris and CHEC, we first threshold, then apply non-maximum suppression.

In simple low-texture environments (such as \texttt{shapes}) each method performs well finding similar points.
In contrast, in high-texture environments (\texttt{dynamic}), state-of-the-art event-based detectors tend to find many spurious corners. 
The identification of too many corners is as much a problem for image processing pipelines as the failure of an algorithm to detect good corners.
The CHEC detector demonstrates a very similar response to the classical Harris algorithm on large sections of the image.
Points identified appear to be well correlated with visual corners in the image and correlate well with the corners identified by the classical Harris corner detector.
We emphasise that the two algorithms use completely separate data, the CHEC algorithm uses the event stream while the classical Harris algorithm is using the conventional frame output of the DAVIS camera. 

On areas of high dynamic range, such as under the chair, on the first authors face, and in the top right of the image, the classical Harris algorithm is unable to extract sufficient contrast to generate an effective corner response while the CHEC algorithm functions effectively.
Furthermore, the output of the CHEC algorithm will not suffer from image blur and can be computed asynchronously in real time, providing an ideal front end corner detector for real-world robotic systems.

\section{CONCLUSION} \label{sec:conclusion}

We have introduced a method to compute spatial convolutions for event cameras.
A key feature is the continuous-time internal state that encodes convolved image information and allows asynchronous, event-driven, incremental updates.
We extend the concept of an internal state to a Harris corner-response state, and demonstrate corner detection (CHEC) without requiring intensity.
We believe there are many exciting possibilities in this direction, including alternative feature states, continuous-time optical flow state, and application of event-based convolutions to convolutional neural networks.

\bibliographystyle{IEEEtran}
\bibliography{all,extra_references}

\end{document}